\documentclass{article} 
\usepackage{iclr2025_conference,times}
\iclrfinalcopy

\usepackage{amsmath,amsfonts,bm}

\def\eqref#1{equation~\ref{#1}}

\def\1{\bm{1}}

\DeclareMathAlphabet{\mathsfit}{\encodingdefault}{\sfdefault}{m}{sl}
\SetMathAlphabet{\mathsfit}{bold}{\encodingdefault}{\sfdefault}{bx}{n}

\usepackage{hyperref}
\usepackage{url}
\usepackage{amsmath}
\usepackage{graphicx}
\usepackage{wrapfig}
\usepackage{multirow}
\usepackage{dsfont}
\usepackage{subcaption}
\usepackage{float}
\usepackage{booktabs}
\usepackage{makecell}

\newtheorem{definition}{Definition}

\title{Fine-Tuning Attention Modules Only: \\ Enhancing Weight Disentanglement in Task Arithmetic}

\author{%
  Ruochen Jin*$^{1,2}$,
  Bojian Hou*$^{2}$, 
  Jiancong Xiao*$^{2}$,
  Weijie Su$^{2}$,
  \textbf{and Li Shen$^{2}$}\\
  $^1$ East China Normal University, Shanghai, China, \\
  $^{2}$  University of Pennsylvania, Philadelphia, PA, USA\\
\texttt{\{kyrie.jin, li.shen\}@pennmedicine.upenn.edu}
}  

\begin{document}

\maketitle


\begin{abstract}
In recent years, \textit{task arithmetic} has garnered increasing attention. This approach edits pre-trained models directly in weight space by combining the fine-tuned weights of various tasks into a \textit{unified model}. Its efficiency and cost-effectiveness stem from its training-free combination, contrasting with traditional methods that require model training on large datasets for multiple tasks. However, applying such a unified model to individual tasks can lead to interference from other tasks (lack of \textit{weight disentanglement}). To address this issue, Neural Tangent Kernel (NTK) linearization has been employed to leverage a ``kernel behavior'', facilitating weight disentanglement and mitigating adverse effects from unrelated tasks. Despite its benefits, NTK linearization presents drawbacks, including doubled training costs, as well as reduced performance of individual models.
To tackle this problem, we propose a simple yet effective and efficient method that is to finetune the attention modules only in the Transformer. Our study reveals that the attention modules exhibit kernel behavior, and fine-tuning the attention modules only significantly improves weight disentanglement. 
To further understand how our method improves the weight disentanglement of task arithmetic, we present a comprehensive study of task arithmetic by differentiating the role of the representation module and task-specific module. In particular, we find that the representation module plays an important role in improving weight disentanglement whereas the task-specific modules such as the classification heads can degenerate the weight disentanglement performance.  \footnote{The code is available at \url{https://github.com/kyrie-23/task_arithmetic_tangent}} 



\end{abstract}

\section{Introduction}

The emergence of large pre-trained models in the open-source community has significantly expanded the potential to enhance performance on downstream tasks \citep{ilharco2023editing,ilharco2022patching,zhuang2020comprehensive}, align with human preferences \citep{lu2022quark,ouyang2022training,ribeiro2022adaptive,glaese2022improving,xiao2024algorithmic}, and improve robustness \citep{hou2017learning,ortiz-jimenez2021optimism,santurkar2021editing,tancik2020fourier}. However, traditional methods often involve expensive joint fine-tuning across multiple tasks \citep{zhuang2020comprehensive} and rely heavily on human feedback \citep{ouyang2022training}, which limits their scalability and broad adoption. Moreover, optimizing performance for specific downstream tasks usually compromises the model's initial pre-training performance or zero-shot accuracy \citep{french1999catastrophic,mccloskey1989catastrophic}.

In light of these challenges, the necessity of task arithmetic in multitask learning has become increasingly evident. \textit{Task arithmetic} offers a cost-effective and efficient alternative by enabling training-free combinations in the weight space of pre-trained models without sacrificing the model's original capabilities \citep{ilharco2023editing}. Central to this approach is the concept of a \emph{task vector}, which represents a set of weight adjustments specifically calibrated for a given task through fine-tuning, obtained by subtracting the task-specific weights from the original pre-trained weights \citep{ilharco2023editing}. Each task vector encodes a unique representational signature tailored to a particular task. 

\begin{figure}
    \centering
    \includegraphics[width=0.9\textwidth]{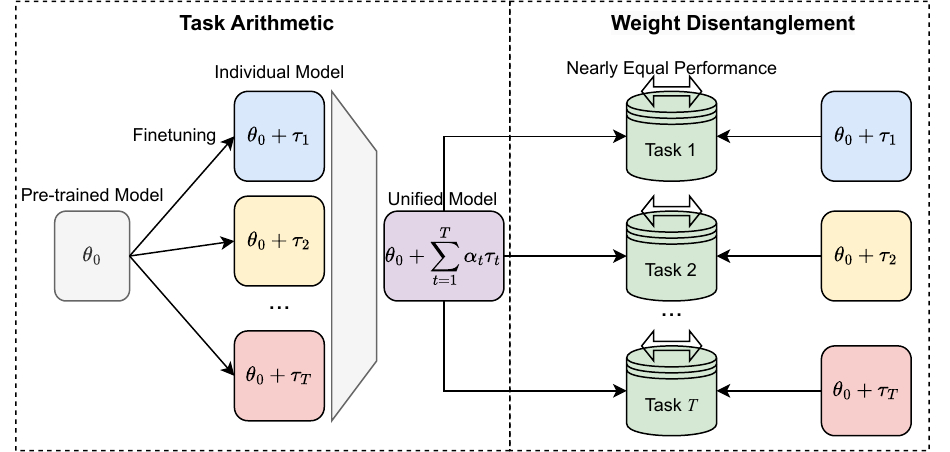}
    \caption{\textbf{Illustration of the concepts of task arithmetic and weight disentanglement.} On the left-hand side, in task arithmetic, we first finetune the pre-trained model $\theta_0$ and get the finetuned individual model $\theta_0+{\alpha_t}\tau_t$ where $\tau_t$ is the $t$th task vector. We eventually obtain the unified model by adding all the task vectors to the pre-trained model: $\theta_0+\sum_{t=1}^T{\alpha_t}\tau_t$. 
    On the right-hand side, weight disentanglement means that the prediction of the unified model on a specific task will not be affected by other tasks.}
    \label{fig:task arithmetic}
\vspace{-10pt}
\end{figure}

As illustrated in Figure \ref{fig:task arithmetic} (left),
$\theta_0$ is the pretrained model and $\tau_t,\ t=1,\cdots,T$ is the $t$th task vector. The \textit{individual model} on each task is derived by $\theta_0+{\alpha_t}\tau_t$. Task arithmetic is to add all the task vectors to the pre-trained model to obtain the \textit{unified model} $\theta_0+\sum_{t=1}^T{\alpha_t}\tau_t$. 
A main goal in task arithmetic is to achieve weight disentanglement as shown in Figure~\ref{fig:task arithmetic} (right) which means the prediction performance of the unified model such as the accuracy on a specific task will not be affected by other tasks. In other words, the unified model has nearly equal performance with the individual model. 
The reason that we aim to achieve weight disentanglement is because task arithmetic primarily focuses on competing tasks rather than synergistic ones~\citep{ilharco2023editing}, as the challenge lies in balancing and optimizing performance across tasks that may have conflicting objectives or require different model behaviors. This emphasis on competing tasks is crucial for developing robust multi-task models that can effectively handle a diverse range of applications without performance degradation.

\begin{table}[t]
\setlength{\tabcolsep}{4pt}
    \centering
    \small
        \caption{
        \textbf{Taks arithmetic performance comparison between different methods.} The task arithmetic performance (the average accuracy of the unified model over all the tasks) of our method outperforms the state-of-the-art due to both good performance of individual models and good kernel behavior (weight disentanglement).}
    \begin{tabular}[width=\columnwidth]{l|ccc}
    \toprule
    & \makecell{Performance of \\ Individual Models} & \makecell{Kernel Behavior \\ (Weight Disentanglement)} & \makecell{Task Arithmetic \\ Performance}\\
    \midrule
    \citet{ilharco2023editing}  & \checkmark&-& 70.00\%\\
        \citet{ortiz-jimenez2024task} & -&\checkmark&76.26\%\\ \midrule
        \textbf{Ours}&\checkmark&\checkmark&78.37\%\\
        \bottomrule
    \end{tabular}
    \label{tab:compare}
\end{table}

However, weight disentanglement remains the most formidable challenge in task arithmetic. Recent research \citep{ortiz-jimenez2024task} has shown that constraining models to fine-tune within the \textit{tangent space} significantly improves weight disentanglement, thanks to the inherent \textit{kernel behavior} of models during early fine-tuning stages. This kernel behavior, formalized by the Neural Tangent Kernel (NTK) theory \citep{jacot2018neural}, refers to neural networks updating primarily around pre-trained parameter initializations. While NTK linearization is effective, it compromises the performance of individual models and demands two to three times more computational resources, conflicting with task arithmetic's original efficiency goals.

\begin{figure}
    \centering
    \includegraphics[width=\columnwidth]{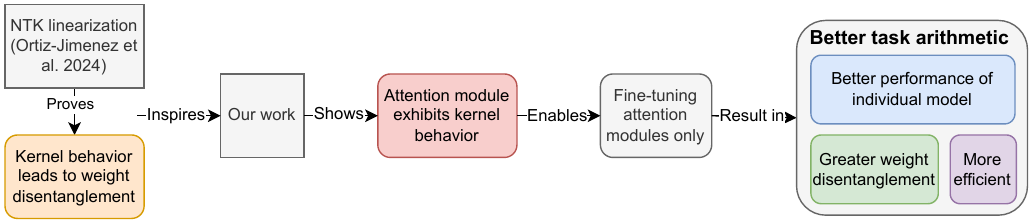}
    \caption{Logic flow of our work.}
    \label{fig:logic flow}
\end{figure}

To tackle this problem, we hypothesize that a combination of non-linear fine-tuning and weight disentanglement is necessary. This hypothesis motivates us to investigate the following question: Does there exist a sub-module within neural networks where non-linear fine-tuning can exhibit weight disentanglement?
We demonstrate that attention modules in transformer models exhibit kernel behavior (see Figure \ref{fig:post-hoc}) which is crucial for weight disentanglement. This finding is supported by empirical evidence using post-hoc linearization~\citep{ortiz-jimenez2024task}, a technique that approximates the change in network output after training using a first-order Taylor expansion. 
Building upon the insights above, we propose to fine-tune the attention modules only, which results in improving the performance of individual models and efficiency while maintaining the weight disentanglement. This logic flow is illustrated in Figure~\ref{fig:logic flow}.


By focusing on fine-tuning only the attention modules, our approach significantly improves weight disentanglement and accuracy of the unified model compared to non-linear fine-tuning and NTK linearization (see Table \ref{tab:addition}), while substantially reducing computational burden and memory usage. This method offers a balanced alternative that maintains strong performance of individual models while enhancing weight disentanglement capabilities, providing a practical solution for improving task arithmetic performance without sacrificing efficiency or accuracy. 

To further understand how our method improves the weight disentanglement of task arithmetic, we present a study by differentiating the role of the representation module and task-specific module, while existing literature \citep{ortiz-jimenez2024task} formulated task arithmetic using a single model without clearly differentiating them. We conduct a comprehensive study of task arithmetic on pre-trained Vision Transformer (ViT) models like the Contrastive Language-Image Pre-Training (CLIP) model \citep{radford2021learning}, providing new insights into its fundamental mechanisms and proposing novel methods to improve the performance of pre-trained models through task arithmetic. 
Specifically, we illustrate that the representation module plays an important role in improving weight disentanglement whereas this has been constrained by task-specific modules, such as classification heads.

In particular, our main contributions are as follows:
\begin{itemize}
\item We propose a simple yet effective and efficient method that only fine-tunes attention modules, which improves weight disentanglement and the average accuracy of the unified model on all the tasks up to 2.38\% improvement compared to the state-of-the-art methods and 8.37\% over the non-linear baseline on several vision-language benchmarks.
\item We demonstrate that the attention module exhibits kernel behavior, suggesting that focusing on fine-tuning these modules could enhance the weight disentanglement capabilities in task arithmetic while maintaining efficiency.
\item We reformulate the architecture of task arithmetic by separating the representation module from task-specific modules, revealing that while weight disentanglement mostly comes from the representation module, the effectiveness of task arithmetic is constrained by task-specific components like classification heads.

\end{itemize}

\section{Preliminaries: Task Arithmetic and Weight Disentanglement}
\label{sec:pre}

We begin by introducing the necessary mathematical notations. Let \( F: \mathcal{X} \times \Theta \rightarrow \mathcal{Y} \) be a neural network taking inputs \( x \in \mathcal{X} \) and parameterized by a set of weights \( \vartheta \in \Theta \), which consists of a representation module $f(\cdot;\theta)$ and a task-specific module $g(\cdot;\omega)$ where $\vartheta=\{\theta,\omega\}$. We assume \( \mathcal{X} \subseteq \mathbb{R}^d \), \( \Theta \subseteq \mathbb{R}^m \) and \( \mathcal{Y} \subseteq \mathbb{R}^c \). Consider \( T \) tasks, with every task \( t \) consisting of a triplet \( (D_t, \mu_t, F_t^*) \), where \( D_t \subseteq \mathcal{X} \) is a data support (e.g., ImageNet \citep{deng2009imagenet} images), \( \mu_t \) an input distribution such that \( \text{supp}(\mu_t) = D_t \), and \( F_t^* : D_t \rightarrow \mathcal{Y} \) a target function (e.g., labels). In practice, each task is identified with a training set \( \{(x_v, F_t^*(x_v))\}_{v \in [n]} \) where $F_t^*(x_v)=g(f(x_v;\theta_t^*);\omega_t)$ with \( x \sim \mu_t \), that is used to fine-tune the representation modules starting from the pre-trained weights \( \theta_0 \) and to obtain the fine-tuned weights \( \theta_t^* \), while the task-specific modules are fixed at $\omega_t$.

Having established this context, we can now introduce the key concepts of this work: task vectors and task arithmetic.

\begin{definition}[Task Vector and Task Arithmetic]
\label{def:task}
Let $\theta_0$ denote the parameters of the pre-trained representation modules and $\theta_t$ denote the parameters after fine-tuning on task $t$. The task vector $\tau_t$ for task $t$ is defined as: $\tau_t = \theta_t - \theta_0$. Task arithmetic is an operation of adapting a pre-trained model to $T$ different tasks by modifying the pre-trained parameters $\theta_0$ to the unified parameters $\theta_\textnormal{unified}$ as follows:
\[ \theta_\textnormal{unified} = \theta_0 + \sum_{t=1}^T \alpha_t \tau_t, \]
where $\alpha_t$ are scalar coefficients.
\end{definition}
To distinguish from the concept of a \textit{unified model}, we refer to the model $f_{\theta_t}$ fine-tuned on a specific task $T$ as an \textit{individual model}. It is worth noting that adding a single task vector $\tau_t$ to a pre-trained model with a coefficient $\alpha = 1$ yields a model equivalent to this individual model $f_{\theta_t}$.

\paragraph{Non-linear Fine-Tuning.} The initial idea of task arithmetic was introduced by \citet{ilharco2023editing}. They demonstrated that performance of the unified model on multiple tasks could be improved simultaneously through task arithmetic. However, 
the performance of the model still lagged behind that of individual models specifically fine-tuned for particular tasks. In the remainder of this paper, we refer to this baseline method as \emph{non-linear fine-tuning}, as the task vectors $\tau_t$ are obtained through this approach.

\paragraph{Accuracy Gap.} We define the \emph{accuracy gap} as the difference in accuracy on task $t$ between the unified model and the corresponding individual model. A plausible hypothesis for this accuracy gap is that the task vectors for different tasks exhibit implicit conflicts with one another. To address the limitations of non-linear fine-tuning, \citet{ortiz-jimenez2024task} proposed that \emph{weight disentanglement} is an important and potentially necessary condition for effective task arithmetic.

\begin{definition}[Weight disentanglement]\label{def:weight disentanglement}
   Given $T$ different tasks and their corresponding supports $D = \{D_t\}_{t \in [T]}$. We say a set of task vectors $\mathcal{T} = \{\tau_t\}_{t \in [T]}$ is weight disentangled with respect to a parametric function $f: \mathcal{X} \times \Theta \rightarrow \mathcal{Y}$ and the initial weights $\theta_0$, if 
\begin{align}
f\left( x; \theta_0 + \sum_{t=1}^T \alpha_t \tau_t \right) =  {\sum_{t=1}^T}f(x; \theta_0 + \alpha_t \tau_t) \mathds{1}(x \in D_t) + f(x; \theta_0) \mathds{1}\left( x \notin \bigcup_{t \in [T]} D_t \right).
\label{eq:disentanglement}
\end{align}
where $f$ is the representation module.
\end{definition}
 
\textbf{Remark.} Although the concept of weight disentanglement was originally proposed by \citet{ortiz-jimenez2024task}, our Definition \ref{def:weight disentanglement} differs from theirs in several key aspects. 

Firstly, our definition characterizes weight disentanglement as a property of a set of task vectors w.r.t. the function $f$, whereas it was originally defined as a property of the function $f$ w.r.t. the task vectors. There are two reasons for our reversed formulation. 1) The term ``weight disentanglement'' corresponds to $T$ different weights (i.e., task vectors), rather than $f$. 2) Both \citet{ortiz-jimenez2024task}'s and our approach aims to find better task vectors $\mathcal{T}$, rather than finding better $f$, for task arithmetic.

Secondly, our definition applies specifically to the representation module, whereas the original definition encompasses the entire neural network. This is primarily motivated by the fact that task arithmetic is performed exclusively on the representation module. We will show later in Section \ref{sec:disentanglement} that this definition is well-defined: weight disentanglement emerges from the representation module.

\paragraph{NTK Linearization Fine-tuning.} Inspired by studies of NTK showing that very wide neural networks behave similarly to linear functions around their initialization $\theta_0$, \citet{ortiz-jimenez2021optimism} proposed fine-tuning the task vectors $\tau$ on a post-hoc linear function, denoted as $f_\text{lin}$, of \( f \) at \( \theta_0 \):
\[
f_{\text{lin}}(x; \theta_0 + \tau) = f(x; \theta_0) + \tau^\top \nabla_\theta f(x; \theta_0).
\]
The obtained task vectors are denoted as $\mathcal{T}_\textnormal{lin}$. The authors further demonstrated empirically that this approach yields an important property: $\mathcal{T}_\textnormal{lin}$ is weight disentangled w.r.t. $f_\textnormal{lin}$ at $\theta_0$.

To conclude this section, we define good task arithmetic performance as characterized by three key factors: high accuracy, computational efficiency, and good weight disentanglement.


\section{Task Arithmetic in Attention Modules}
\subsection{Main Challenge of Task Arithmetic}
Based on our previous definition, the performance of the unified model on each task $t$ can be simply decomposed as follows:
\begin{center}
    \emph{Accuracy of unified models = Accuracy of individual models - Accuracy Gap.}
\end{center}
Unfortunately, existing studies reveal a trade-off between these factors. Non-linear fine-tuning can achieve high accuracy for individual models, but the resulting task vectors are less weight disentangled, leading to a larger accuracy gap in the unified model. Conversely, linear fine-tuning guarantees weight disentanglement and thus a lower accuracy gap, but the linearly fine-tuned individual models tend to be less accurate, as linear approximation brings error to the models. This trade-off presents a significant challenge in task arithmetic.

To this aim, we hypothesize that a combination of non-linear fine-tuning and weight disentanglement is necessary. This hypothesis motivates us to investigate the following question: Does there exist a sub-module within neural networks where non-linear fine-tuning can exhibit weight disentanglement? Our investigation reveals that the attention module is a promising candidate.

\subsection{Accuracy Gap: Kernel Behavior and Weight Disentanglement of Attention Module}
\label{sec:kernel}

\begin{wrapfigure}{r}{0.5\textwidth}
\vspace{-0.22in}
    \centering
    \vspace{-10pt}
    \includegraphics[width=0.5\textwidth]{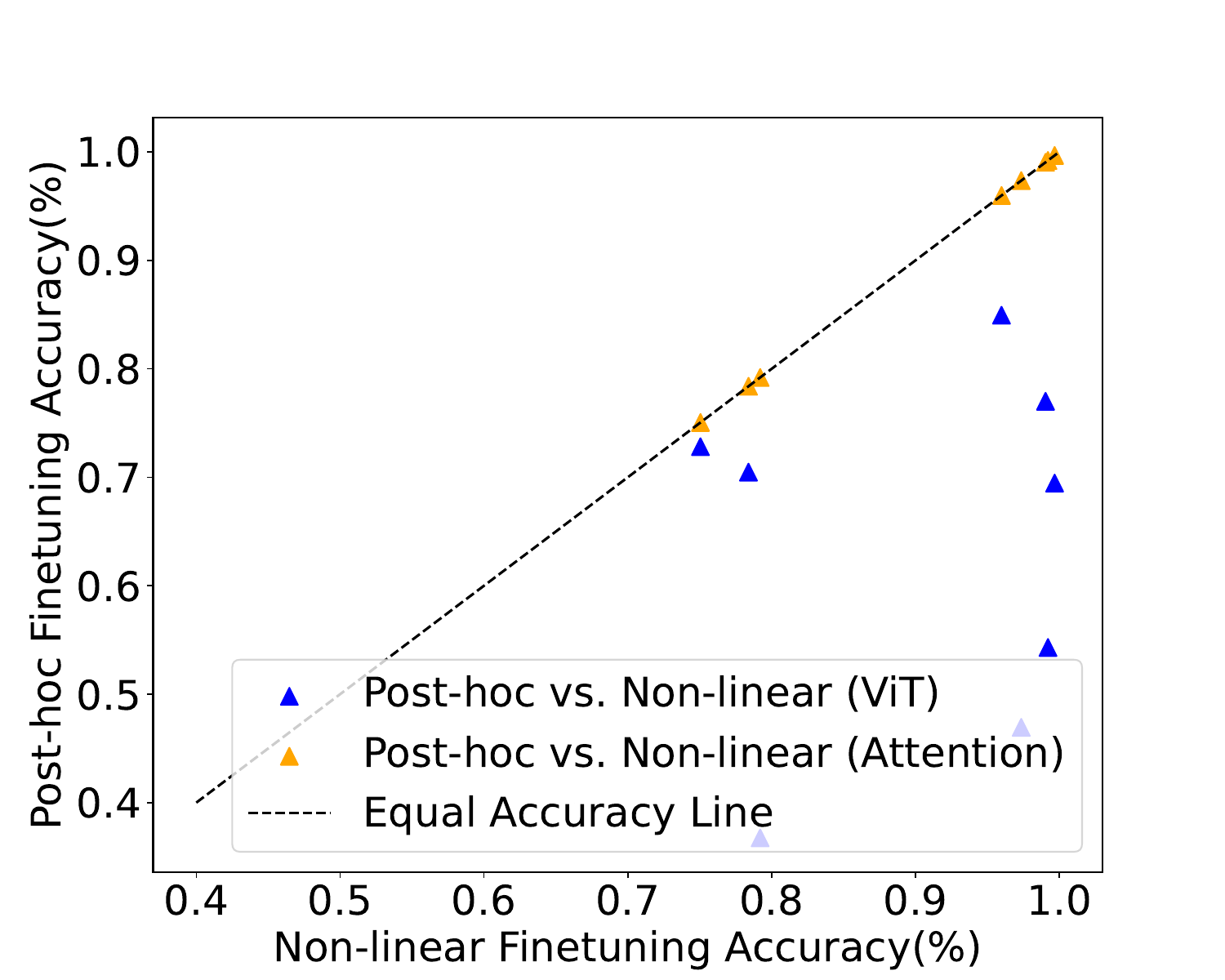}
    \vspace{-10pt}
    \caption{Accuracy of non-linear and post-hoc models by tasks. The diagonal dashed line indicates post-hoc performance meets non-linear. }
    \vspace{-10pt}
    \label{fig:post-hoc}
\end{wrapfigure}
As per the study of NTK studies and the discussion in \cite{ortiz-jimenez2024task}, if a pre-trained network \( f(\cdot; \theta_0) \) exhibits \textit{kernel behavior} during fine-tuning -- that is, if the neural network updates primarily around its pre-trained parameter initialization and can be approximated by its first-order Taylor expansion -- then the resulting task vectors are weight disentangled. Directly examining whether kernel behavior can be challenging; however, it can be approximated by the following test.

\paragraph{Kernel Behavior Test.} Given a function $f$ with initial parameters $\theta_0$, and a task $t$ with task vector $\tau_t$, we define the Kernel Behavior Test as follows. If the equation
\[
f(x; \theta_0 + \tau_t) = f_{\text{lin}}(x; \theta_0+\tau_t),
\]
holds for all $x$ in the dataset of task $t$, we say that $f(\cdot;\theta)$ exhibits kernel behavior during fine-tuning using the given approach on task $t$, or more simply, that the given approach exhibits kernel behavior.

In our experiments, we compare the average accuracy across $T$ tasks of the non-linear function ($f(\cdot)$) and the post-hoc linear function ($f_{\text{lin}}$), fune-tuning on (1) all the parameters and (2) only on the attention modules. Specifically, we fine-tune several CLIP pre-trained ViTs \citep{dosovitskiy2021image} of different sizes following the same setup as \citet{ilharco2023editing} on 8 tasks: Cars \citep{krause2013object}, DTD \citep{cimpoi2014describing}, SUN397 \citep{xiao2016sun}, EuroSAT \citep{helber2019eurosat}, GTSRB \citep{stallkamp2011german}, MNIST \citep{lecun1998mnist}, SVHN \citep{netzer2011reading} and RESISC45 \citep{cheng2017remote}. 

The results in Figure \ref{fig:post-hoc} indicate that the attention module demonstrates kernel behavior. The triangle dots show the comparison of the kernel behavior test between the attention modules (yellow) and the whole models (blue), respectively. The proximity of dots to the diagonal dashed line indicates kernel behavior. The post-hoc of attention module, represented by yellow dots, consistently appears closer to the diagonal dashed line than the whole ViT (blue dots), suggesting superior performance. 

This indicates that attention modules demonstrate stronger kernel behavior compared to the full model, suggesting that focusing on fine-tuning these modules could enhance the weight disentanglement, finally resulting in a low accuracy gap.

\subsection{Accuracy of Individual Models with Fine-Tuning Attention Modules}

\begin{figure}
    \centering
    \includegraphics[width=0.8\columnwidth]{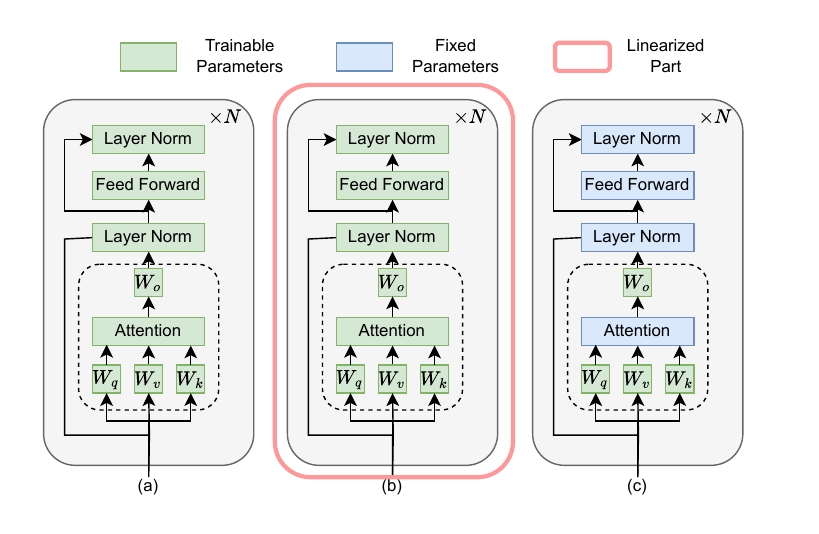}
    \vspace{-0.8cm}
    \caption{\textbf{Three types of fine-tuning paradigms.} (a) Non-linear fine-tuning where all the parameters will be updated. (b) Full-model linearization. (c) Attention modules only fine-tuning where only $W_q$, $W_v$, $W_k$ and $W_o$ will be updated. In this paper, we explore attention modules only fine-tuning.}
    \label{fig:paradigm}
\vspace{-0.5cm}
\end{figure}

Based on the kernel behavior of attention modules, we propose focusing on fine-tuning only the attention modules. The comparison of our fine-tuning paradigms with the non-linear fine-tuning paradigm and the NTK linearization fine-tuning paradigm is demonstrated in Figure \ref{fig:paradigm}. Next, we will show that fine-tuning attention modules also achieves high accuracy for individual models.


\textbf{Non-Linear Advantage.} We will first introduce a crucial concept referred to as \emph{non-linear advantage}. For a given approach, the non-linear advantage is defined as the difference in accuracy of individual models between non-linear fine-tuning and the approach in question. Since non-linear fine-tuning typically achieves the highest accuracy for individual models, the non-linear advantage is always non-negative, i.e., non-linear advantage $\geq 0$.

\begin{wrapfigure}{r}{0.45\textwidth}
\vspace{-0.7cm}
    \centering
    \includegraphics[width=0.45\textwidth]{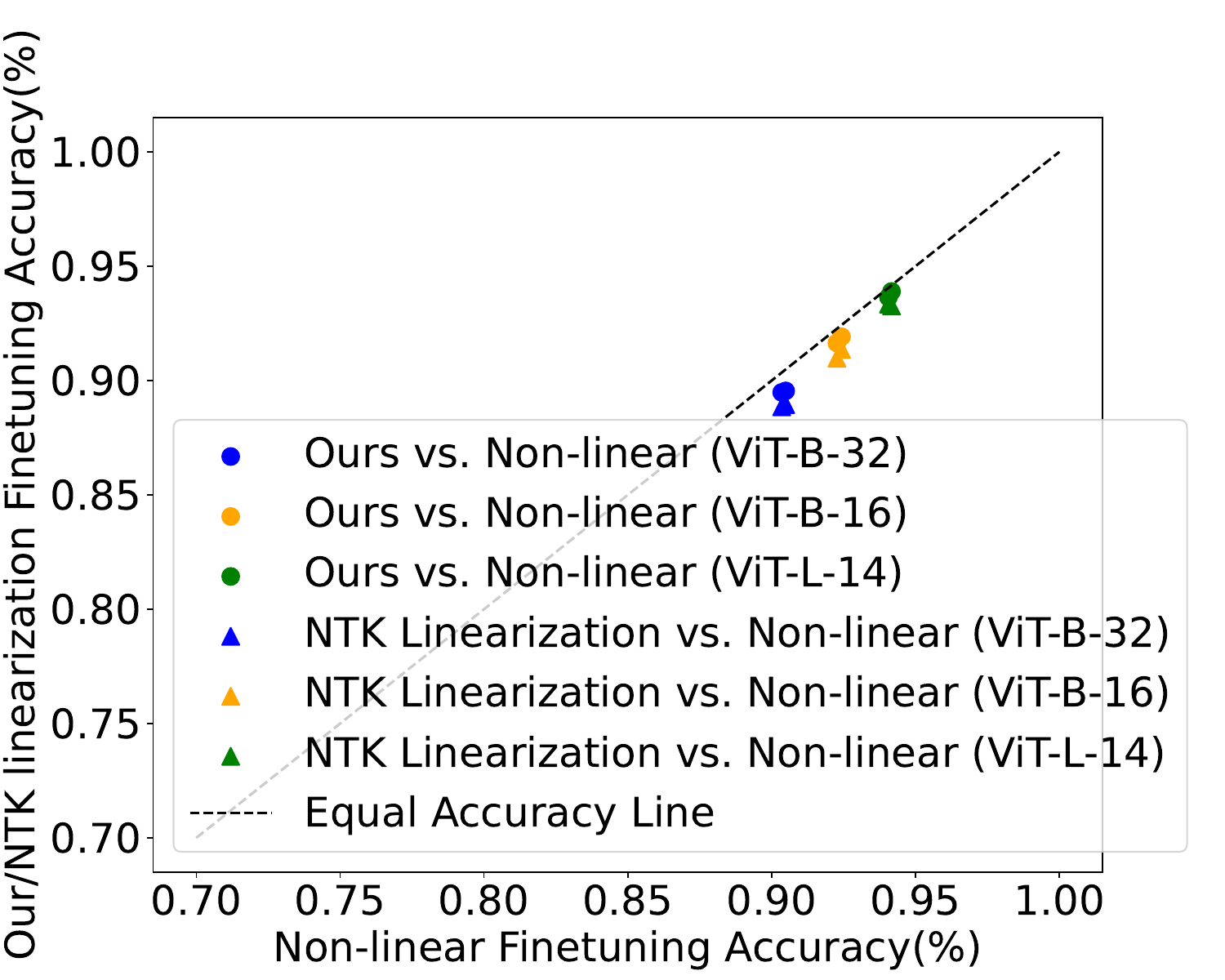}
    \vspace{-10pt}
    \caption{Averaged accuracy of non-linear and linear models. The diagonal dashed line indicates linear fine-tuning performance meets non-linear.}
    \label{fig:single-task}
\vspace{-1.3cm}
\end{wrapfigure}

\textbf{Accuracy of Individual Models.} In Figure \ref{fig:single-task}, we demonstrate that fine-tuning attention modules can reduce the non-linear advantage—essentially improving the accuracy of individual models—compared to NTK linearization fine-tuning\footnote{Please see Appendix \ref{app:fine-tuning} for performance on each task.}. The figure presents two comparisons: 1) Round markers represent the comparison between our method and non-linear fine-tuning. 2) Triangular markers show the comparison between NTK linearization and non-linear fine-tuning.

The proximity of markers to the diagonal dashed line indicates a non-linear advantage equal to zero. Our method, represented by round markers, consistently appears closer to the diagonal dashed line than the NTK linearization (triangular markers), indicating a smaller non-linear advantage.
This visual representation demonstrates that our approach of fine-tuning attention modules achieves performance closer to that of non-linear fine-tuning compared to NTK linearization, thus reducing the non-linear advantage more effectively.

\begin{table}
\centering
\caption{\textbf{Comparison of performance for task arithmetic across various visual models.} This table presents the average accuracy (\%) and normalized accuracy (\%) of various ViTs after incorporating the sum of task vectors from eight different tasks.  We report results for the non-linear fine-tuning and NTK linearized models normalizing performance by their single-task accuracy.}

\begin{tabular}[width=\columnwidth]{lll|ll|ll}
\toprule
\multirow{2}{*}{Method} & \multicolumn{2}{c|}{ViT-B-32}   & \multicolumn{2}{c|}{ViT-B-16}   & \multicolumn{2}{c}{ViT-L-14}    \\
                        & Abs.({\color{blue}$\uparrow$})            & Norm.({\color{blue}$\uparrow$})           & Abs.({\color{blue}$\uparrow$})            & Norm.({\color{blue}$\uparrow$})           & Abs.({\color{blue}$\uparrow$})            & Norm.({\color{blue}$\uparrow$})           \\ \midrule
Pre-trained             & 48.40          & -              & 55.25          & -              & 66.40          & -              \\
Non-linear Fine-tuning             & 70.00          & 77.04          & 74.75          & 80.59          & 84.40          & 89.47  \\        
NTK Linearization                  & 76.26          & 85.82          & 79.01          & 86.32          & 85.53          & 91.44          \\ \midrule
\textbf{Ours}           & \textbf{78.37} & \textbf{87.42} & \textbf{80.44} & \textbf{87.25} & \textbf{87.91} & \textbf{93.66} \\ \bottomrule
\end{tabular}
\label{tab:addition}
\end{table}

\begin{table}
\centering
\setlength{\tabcolsep}{1.5pt}
\caption{{\textbf{Comparison of performance for task arithmetic across different language models.} This table presents the average accuracy (\%) of Flan-T5-base models on GLUE benchmark.}}
\begin{tabular}[width=\columnwidth]{l|cccccccc}
\toprule
{ } & {glue-cola} & {glue-mnli} & {glue-mrpc} & {glue-qqp} & {glue-rte} & {glue-sst2} & {glue-stsb} & {avg} \\ \midrule
\makecell{{Non-linear} {Fine-tuning}} & {79.87}   & {80.94}   & {59.31}   & {82.19}  & {50.54}  & {89.33}   & {70.55}   & {73.25} \\ 
{NTK Linearization}      & {75.93}   & {83.19}   & {76.72}   & {87.88}  & {62.09}  & {92.09}   & {66.51}   & {77.77} \\ \midrule
{\textbf{Ours}}     & {80.63}   & {86.25}   & {86.76}   & {89.69}  & {72.20}  & {93.92}   & {87.67}   & \textbf{{85.30}} \\ 
\bottomrule
\end{tabular}
\label{tab:nlp}
\end{table}

\subsection{Accuracy of Unified Models with Fine-tuning Attention Modules}
We have demonstrated that our method has achieved both high accuracy for individual models and solved the accuracy gap. Then we will validate that our method has achieved high accuracy for unified models in terms of average accuracy and normalized accuracy.

To obtain the task vectors, we use the fine-tuned weights of different ViTs from before and use the same mixing coefficient for all tasks, i.e., $\alpha=\alpha_1=\dots=\alpha_T$ to ensure a fair comparison {with \cite{ortiz-jimenez2024task}}. We provide all the details of this experiment in Appendix \ref{app:experiment}.

\textbf{Normalized Accuracy.}
The normalized accuracy is calculated by the individual accuracy achieved by the model fine-tuned on each task. Mathematically,
\begin{equation*}\footnotesize
\text{Normalized Acc} = \frac{1}{T} \sum_{t=1}^T \frac{ \left[\text{acc} \left( f(x; \theta_0 + \sum_t {\alpha_t}\tau_t) \right) \right]}{ \left[\text{acc} \left( f(x; \theta_0 + {\alpha_t}\tau_t) \right) \right]}.
\end{equation*}
\textbf{Accuracy of Unified Models.} 
We employ the benchmark proposed by \cite{ilharco2023editing} to evaluate the task arithmetic ability of a pre-trained model, which consists of the 8 tasks described in Section \ref{sec:kernel}.
In particular, Table \ref{tab:addition} shows that our method significantly outperforms its non-linear counterparts \citep{ilharco2023editing} and achieves state-of-the-art results on the task addition benchmarks. Our model achieves higher accuracy of the unified model through task addition (up to 2.38\%). Additionally, our method not only outperforms on averaged accuracy but also on normalization accuracy.

\begin{table}[t]
\caption{\textbf{Efficiency comparison in terms of parameters, size, and training time.} The NTK linearization requires two to three times more computational resources and doubles the memory footprint compared to its non-linear counterpart. Our method outperforms NTK linearization in accuracy with only a quarter of the costs.}
\centering
\setlength{\tabcolsep}{3pt}
\begin{tabular}{l|cccc}
\toprule
         & Total Params & Trainable Params & \makecell[c]{Computational \\ Cost (MB)} & Training Time (Min) \\ \midrule
Non-linear Fine-tuning & 222 M        & 222 M            & 891.614             & 17:38               \\
NTK Linearization     & 470 M        & 222 M            & 1,881.926           & 40:59               \\
Ours     & 222 M        & 63.7 M           & 891.614             & 10:15              \\ \bottomrule
\end{tabular}
\label{tab:efficiency}
\end{table}

{For Natual Language Processing (NLP) tasks, we utilize the Flan-T5 \citep{Chung2022Scaling} as our pre-trained language model. For fine-tuning, we employ the Flan-T5-base models on seven tasks derived from the General Language Understanding Evaluation (GLUE) benchmark \citep{Wang2018GLUE} with the same random seed 42 to initialize the models. These tasks are CoLA, MNLI, MRPC, QQP, RTE, SST2, and STSB. We report accuracy for all tasks and an average accuracy in Table \ref{tab:nlp}, our method outperforms both non-linear fine-tuning and NTK linearization in vision and NLP tasks.}

\begin{wraptable}{r}{0.52\textwidth}   
\caption{{\textbf{Ablation study.} Single-task and performance of unified model (\%) on 4 different settings.}}
\label{tab:paradigms}
\begin{tabular}{c|c|cc}
\toprule
\multirow{2}{*}{{Paradigm}} & {Single-task} & \multicolumn{2}{c}{{Multi-task}}  \\
                          & {Accuracy({\color{blue}$\uparrow$})}                       & {Abs.({\color{blue}$\uparrow$})}            & {Norm.({\color{blue}$\uparrow$})}           \\ \midrule
{\textbf{(1)}}              & {\textbf{89.55}}               & {\textbf{78.37}} & {\textbf{87.42}} \\
{(2)}                       & {89.48}                          & {77.71}       & {86.79}   \\ \midrule
{(3)}                       & {88.95}                        & {76.52}          & {86.11}          \\
{(4)}                       & {89.43}                        & {77.80}          & {86.93}          \\ \bottomrule
\end{tabular}
\vspace{-10pt}
\end{wraptable}
\textbf{Ablation Study.}
To investigate whether we can enhance weight disentanglement performance by fine-tuning the Multiple-Layer Perceptron (MLP) modules in addition to attention modules, we conduct an ablation experiment with four paradigms: (1) fine-tuning only attention weights ($Q$, $K$, $V$, and $O$ projections) (ours), (2) fine-tuning attention weights and biases, (3) fine-tuning both attention and MLP weights, and (4) fine-tuning attention and MLP weights along with biases. Remarkably, all four paradigms outperformed NTK linearization in terms of both performance and weight disentanglement, indicating that ViT models exhibit strong kernel behavior within the attention modules and MLP. However, performance varied based on whether bias parameters were fine-tuned, with the best results aligning closely with settings used in LoRA. This suggests that further exploration of these configurations could yield valuable insights into optimizing task arithmetic.




\textbf{Efficiency Comparison.} In addition to superior accuracy, we demonstrate our method is much more efficient than non-linear fine-tuning and NTK linearization in Table \ref{tab:efficiency} due to fine-tuning fewer parameters.

Our fine-tuning method significantly enhances the appeal of task arithmetic for practical applications. By improving the performance of individual models, our approach demonstrates the superiority of task arithmetic in achieving good accuracy of the unified model efficiently. Additionally, we have observed that kernel behavior within attention modules fosters greater task disentanglement. In the subsequent section, we will delve deeper into this concept, exploring its implications and potential for future advancements.

\vspace{-10pt}
\section{Robustness of Task Arithmetic with Respect to Coefficient $\alpha$}
\label{sec:disentanglement}
In previous sections, we discuss how to find a good $\mathcal{T}$ in task arithmetic. Yet the robustness of task arithmetic (i.e., the effect of $\alpha$) on weight disentanglement has not been explored. To this end, in the following section, 
we first propose a metric to evaluate the weight disentanglement performance called ``disentanglement error'' on the representation module and prove that weight disentanglement emerges from the representation module. Then, we illustrate that our method has great weight disentanglement for a wider choice of $\alpha$ for both representation and classification modules, which demonstrates the robustness of the task arithmetic of our method.

\subsection{Weight Disentanglement Emerges From Representation Module}

To explore the robustness of task arithmetic and its effect on weight disentanglement, we propose a metric called ``disentanglement error'' to evaluate weight disentanglement performance. Unlike previous work that focused on task-specific modules, we investigate whether the representation module can satisfy Definition \ref{def:weight disentanglement} (Weight Disentanglement) without relying on task-specific components.

Our hypothesis is that pre-trained models can demonstrate task arithmetic properties independently of downstream tasks, maintaining consistent representations through task arithmetic. By focusing on the representation module alone, we aim to show that the inherent properties of pre-trained models are sufficient to support task arithmetic, potentially simplifying the process and broadening its applicability.



\begin{figure}
    \centering
    \includegraphics[width=\columnwidth]{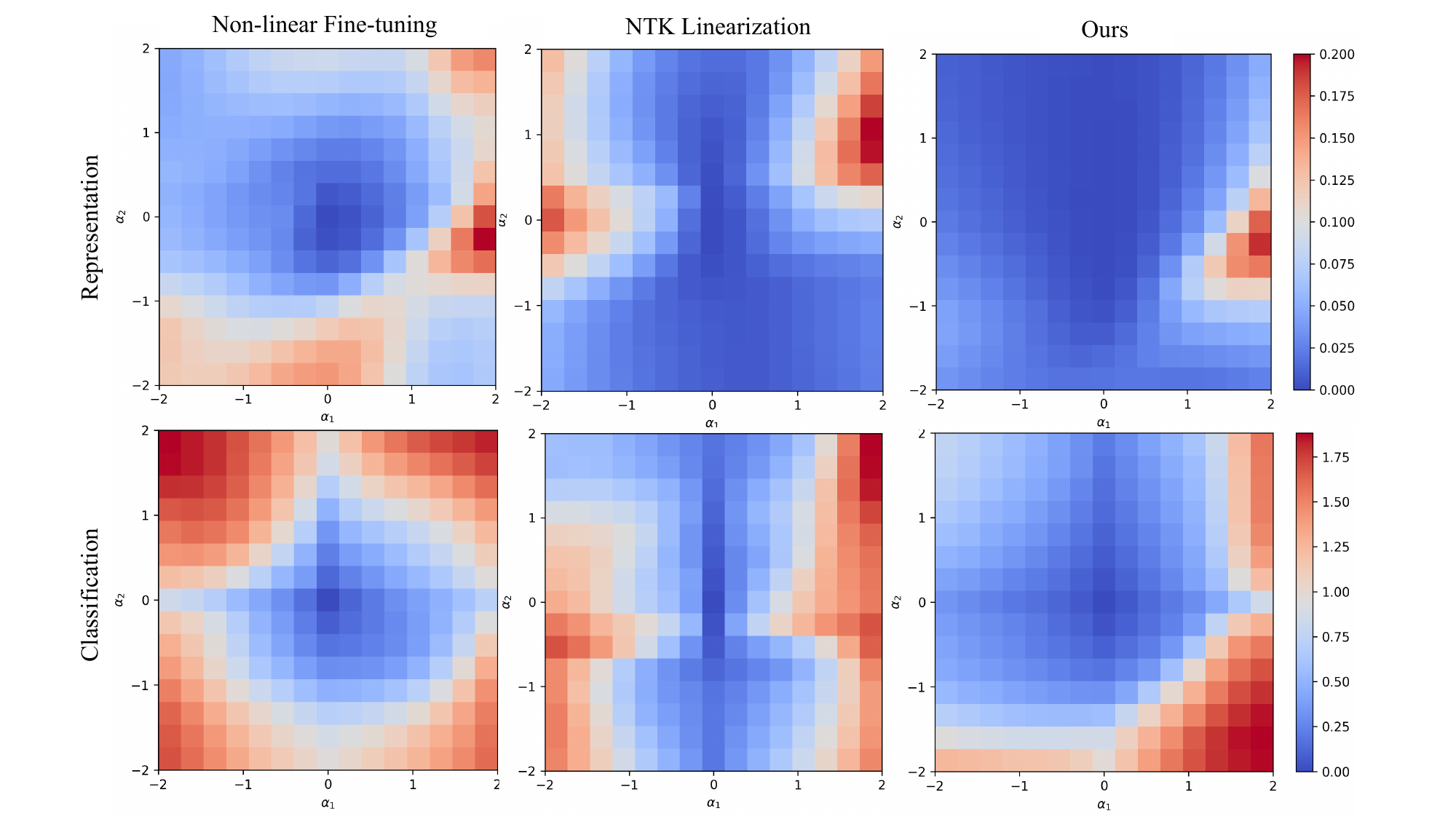}
    \vspace{-0.5cm}
    \caption{\label{fig:disentanglement}\textbf{Visualization of weight disentanglement.} The heatmaps show the disentanglement error $\xi(\alpha_1, \alpha_2)$ (see Eq. (\ref{eq:disen_error})) of a single representation module CLIP ViT-B/32 (top) and a combination of representation module and classification module on DTD - SUN397 task pair. We use prediction error for classification task as Ortiz-Jimenez did \citep{ortiz-jimenez2024task}. Three fine-tuning paradigms are used from left to right: non-linear fine-tuning, NTK linearization, and ours. The blue regions denote areas of the weight space where weight disentanglement is stronger.}
\vspace{-10pt}
\end{figure}

To visualize the level of weight disentanglement, we measure the discrepancy with Eq. (\ref{eq:disentanglement}) using the disentanglement error {\citep{ortiz-jimenez2024task}}:
\begin{align}
\label{eq:disen_error}
    \xi(\alpha_1, \alpha_2) = \sum_{t=1}^{2} \mathbb{E}_{x \sim \mu_t} \left[ \text{dist} \left( f(x; \theta_0 + \alpha_t \tau_t), f(x; \theta_0 + \alpha_1 \tau_1 + \alpha_2 \tau_2) \right) \right],
\end{align}
where ``dist'' denotes any distance metric between output vectors. As we are dealing with representation distributions, in what follows we use the Kullback–Leibler divergence as the distance metric. In general, the smaller the value of \(\xi(\alpha_1, \alpha_2)\) the more weight disentangled a model is at \((\alpha_1, \alpha_2)\).
\vspace{-5pt}
\subsection{Weight Disentanglement Results}
Figure \ref{fig:disentanglement} displays the disentanglement error of a CLIP ViT-B/32 model concerning several task vector pairs from different fine-tuning paradigms. We observe a minimal disentanglement error within a small region surrounding $\theta_0$, which enables task arithmetic. 
Different from disentanglement error at downstream tasks, it remains relatively small even for $\alpha_1,\alpha_2>1$, which indicates the power of task arithmetic has been limited by the performance of task-specific modules (classification head).

\textbf{Disentanglement Error  Comparison.} Our method demonstrates greater weight disentanglement than its counterparts, as evidenced by the more extensive regions with low disentanglement errors in Figure \ref{fig:disentanglement} (right). This explains the higher normalized accuracy achieved (cf. Table \ref{tab:addition}) when fine-tuning attention modules only. The combination of greater weight disentanglement and better performance of individual models results in higher performance of the unified model.

These results demonstrate that our method achieves great weight disentanglement for a wider choice of $\alpha$ for both representation and classification modules, illustrating the robustness of task arithmetic in our approach.

\vspace{-10pt}
\section{Related Work}

{Existing model merging techniques can be broadly categorized into two main types \citep{Yang2024}: (i) Pre-Merging Methods: These methods focus on enhancing the conditions necessary for effective model merging by optimizing the fine-tuning process of individual models. (ii) During Merging Methods: These approaches address task conflicts and interference through various strategies before executing the parameter merging operations.}

{\textbf{Pre-Merging Methods.}
To establish better conditions for model merging, a significant body of work has focused on the fine-tuning processes of independent models. For instance, \cite{ortiz-jimenez2024task} propose fine-tuning within the tangent space of the pre-trained model, leveraging the NTK framework to enhance performance during the fine-tuning stage. However, fine-tuning all parameters in the linearized model can be computationally intensive compared to nonlinear fine-tuning. To mitigate this issue, some studies advocate for selectively linearizing only certain layers; for example, \cite{tang2023parameter} suggest partially linearizing Adapter modules prior to merging them. Additionally, TAFT \citep{Liu2024} introduces an efficient linearization method specifically for Transformer architectures, deriving closed-form linearized solutions that facilitate smoother integration of models. Overall, fine-tuning in the tangent space aids in disentangling both input and weight spaces, thereby reducing potential interference during subsequent model merging.}

{\textbf{During Merging Methods.}
In the context of multi-task learning (MTL), model merging can be effectively achieved by employing various strategies to resolve task conflicts and perform parameter merging operations. Traditional methods often involve averaging or combining weights from multiple models to create a unified system, as demonstrated in prior works \citep{garipov2018loss,ilharco2023editing,wortsman2022model}. However, these basic merging techniques frequently underperform, particularly when tasks interfere with one another. Advanced methods have been developed to address this challenge by incorporating weighted-based strategies that assign different importance levels to task vectors during merging \citep{matena2021merging,ainsworth2023git,stoica2023zipit,yang2023adamerging}. Furthermore, some approaches transform models into sparse subspaces before merging, effectively mitigating task interference and allowing for the removal of insignificant neurons from individual models while enabling the combination of multiple sparse models within a parameter subspace \citep{yadav2023resolving,Tam2023,li2023deep,zhang2023composing,Huh2024,Huang2024}. This innovative perspective opens new avenues for model merging, enhancing overall performance and flexibility in multi-task applications.}

{Our method falls into the Pre-Merging category, focusing on the fine-tuning process and achieving superior performance in task arithmetic with high efficiency.}

\vspace{-10pt}
\section{Discussion}

\vspace{-5pt}
In this work, we conducted a comprehensive analysis of task arithmetic in deep neural networks, uncovering its fundamental mechanisms and enhancing its performance. Our findings reveal that attention modules exhibit kernel behavior, leading to improved weight disentanglement when fine-tuned exclusively, without compromising individual accuracy or efficiency. Crucially, we demonstrated that weight disentanglement emerges primarily from the representation module, while task-specific modules can limit the effectiveness of task arithmetic. This insight opens up new possibilities for applying task arithmetic in unsupervised learning scenarios and broadens its potential applications.

While our approach significantly advances the field of task arithmetic, several limitations and opportunities for future research remain. Current task vectors are constrained to models with identical architectures and initializations due to their reliance on element-wise weight operations. Future studies could explore integrating task arithmetic with partial fine-tuning techniques, focusing on varying numbers of attention blocks. Additionally, investigating the relationship between the sparsity of attention modules and their kernel behavior may provide insights into learnable tasks. Understanding the nuanced impact of fine-tuning bias on model performance and weight disentanglement also presents an important avenue for future research. These investigations could lead to more robust and efficient methods for adapting pre-trained models to various tasks, significantly enhancing their applicability and effectiveness in real-world scenarios.

\bibliography{arithmetic}

\begin{thebibliography}{46}
\providecommand{\natexlab}[1]{#1}
\providecommand{\url}[1]{\texttt{#1}}
\expandafter\ifx\csname urlstyle\endcsname\relax
  \providecommand{\doi}[1]{doi: #1}\else
  \providecommand{\doi}{doi: \begingroup \urlstyle{rm}\Url}\fi

\bibitem[Ainsworth et~al.(2023)Ainsworth, Hayase, and Srinivasa]{ainsworth2023git}
Samuel~K Ainsworth, Jonathan Hayase, and Siddhartha Srinivasa.
\newblock Git re-basin: Merging models modulo permutation symmetries.
\newblock In \emph{International Conference on Learning Representations (ICLR)}, 2023.
\newblock URL \url{https://arxiv.org/abs/2209.04836}.

\bibitem[Cheng et~al.(2017)Cheng, Han, and Lu]{cheng2017remote}
Gong Cheng, Junwei Han, and Xiaoqiang Lu.
\newblock Remote sensing image scene classification: Benchmark and state of the art.
\newblock \emph{Proceedings of the IEEE}, 2017.
\newblock URL \url{https://ieeexplore.ieee.org/document/7891544}.

\bibitem[Cherti et~al.(2023)Cherti, Beaumont, Wightman, Wortsman, Ilharco, Gordon, Schuhmann, Schmidt, and Jitsev]{Cherti_2023_CVPR}
Mehdi Cherti, Romain Beaumont, Ross Wightman, Mitchell Wortsman, Gabriel Ilharco, Cade Gordon, Christoph Schuhmann, Ludwig Schmidt, and Jenia Jitsev.
\newblock Reproducible scaling laws for contrastive language-image learning.
\newblock In \emph{Proceedings of the IEEE/CVF Conference on Computer Vision and Pattern Recognition (CVPR)}, pp.\  2818--2829, June 2023.

\bibitem[Chung et~al.(2022)Chung, Hou, Longpre, Zoph, Tay, Fedus, Li, Wang, Dehghani, Brahma, Webson, Gu, Dai, Suzgun, Chen, Chowdhery, Castro-Ros, Pellat, Robinson, Valter, Narang, Mishra, Yu, Zhao, Huang, Dai, Yu, Petrov, Chi, Dean, Devlin, Roberts, Zhou, Le, and Wei]{Chung2022Scaling}
Hyung~Won Chung, Le~Hou, Shayne Longpre, Barret Zoph, Yi~Tay, William Fedus, Yunxuan Li, Xuezhi Wang, Mostafa Dehghani, Siddhartha Brahma, Albert Webson, Shixiang~Shane Gu, Zhuyun Dai, Mirac Suzgun, Xinyun Chen, Aakanksha Chowdhery, Alex Castro-Ros, Marie Pellat, Kevin Robinson, Dasha Valter, Sharan Narang, Gaurav Mishra, Adams Yu, Vincent Zhao, Yanping Huang, Andrew Dai, Hongkun Yu, Slav Petrov, Ed~H. Chi, Jeff Dean, Jacob Devlin, Adam Roberts, Denny Zhou, Quoc~V. Le, and Jason Wei.
\newblock Scaling instruction-finetuned language models.
\newblock \emph{arXiv preprint arXiv:2210.11416}, December 2022.
\newblock URL \url{http://arxiv.org/abs/2210.11416}.

\bibitem[Cimpoi et~al.(2014)Cimpoi, Maji, Kokkinos, Mohamed, and Vedaldi]{cimpoi2014describing}
Mircea Cimpoi, Subhransu Maji, Iasonas Kokkinos, Sammy Mohamed, and Andrea Vedaldi.
\newblock Describing textures in the wild.
\newblock In \emph{IEEE Conference on Computer Vision and Pattern Recognition (CVPR)}, 2014.
\newblock URL \url{https://openaccess.thecvf.com/content_cvpr_2014/html/Cimpoi_Describing_Textures_in_2014_CVPR_paper.html}.

\bibitem[Deng et~al.(2009)Deng, Dong, Socher, Li, Li, and Fei-Fei]{deng2009imagenet}
Jia Deng, Wei Dong, Richard Socher, Li-Jia Li, Kai Li, and Li~Fei-Fei.
\newblock Imagenet: A large-scale hierarchical image database.
\newblock In \emph{IEEE Conference on Computer Vision and Pattern Recognition (CVPR)}, 2009.
\newblock URL \url{https://ieeexplore.ieee.org/abstract/document/5206848}.

\bibitem[Dosovitskiy et~al.(2021)Dosovitskiy, Beyer, Kolesnikov, Weissenborn, Zhai, Unterthiner, Dehghani, Minderer, Heigold, Gelly, Uszkoreit, and Houlsby]{dosovitskiy2021image}
Alexey Dosovitskiy, Lucas Beyer, Alexander Kolesnikov, Dirk Weissenborn, Xiaohua Zhai, Thomas Unterthiner, Mostafa Dehghani, Matthias Minderer, Georg Heigold, Sylvain Gelly, Jakob Uszkoreit, and Neil Houlsby.
\newblock An image is worth 16x16 words: Transformers for image recognition at scale.
\newblock In \emph{International Conference on Learning Representations (ICLR)}, 2021.
\newblock URL \url{https://openreview.net/forum?id=YicbFdNTTy}.

\bibitem[French(1999)]{french1999catastrophic}
Robert~M French.
\newblock Catastrophic forgetting in connectionist networks.
\newblock \emph{Trends in Cognitive Sciences}, 1999.
\newblock URL \url{https://www.sciencedirect.com/science/article/pii/S1364661399012942}.

\bibitem[Garipov et~al.(2018)Garipov, Izmailov, Podoprikhin, Vetrov, and Wilson]{garipov2018loss}
Timur Garipov, Pavel Izmailov, Dmitrii Podoprikhin, Dmitry Vetrov, and Andrew~Gordon Wilson.
\newblock Loss surfaces, mode connectivity, and fast ensembling of dnns.
\newblock In \emph{Conference on Uncertainty in Artificial Intelligence (UAI)}, 2018.
\newblock URL \url{https://arxiv.org/abs/1802.10026}.

\bibitem[Glaese et~al.(2022)Glaese, McAleese, Trebacz, Aslanides, Firoiu, Ewalds, Rauh, Weidinger, Chadwick, Thacker, Campbell-Gillingham, Uesato, Huang, Comanescu, Yang, See, Dathathri, Greig, Chen, Fritz, Sanchez~Elias, Green, Mokra, Fernando, Wu, Foley, Young, Gabriel, Isaac, Mellor, Hassabis, Kavukcuoglu, Hendricks, and Irving]{glaese2022improving}
Amelia Glaese, Nat McAleese, Maja Trebacz, John Aslanides, Vlad Firoiu, Timo Ewalds, Maribeth Rauh, Laura Weidinger, Martin Chadwick, Phoebe Thacker, Lucy Campbell-Gillingham, Jonathan Uesato, Po-Sen Huang, Ramona Comanescu, Fan Yang, Abigail See, Sumanth Dathathri, Rory Greig, Charlie Chen, Doug Fritz, Jaume Sanchez~Elias, Richard Green, Sona Mokra, Nicholas Fernando, Boxi Wu, Rachel Foley, Susannah Young, Iason Gabriel, William Isaac, John Mellor, Demis Hassabis, Koray Kavukcuoglu, Lisa~Anne Hendricks, and Geoffrey Irving.
\newblock Improving alignment of dialogue agents via targeted human judgements.
\newblock \url{https://www.deepmind.com/blog/building-safer-dialogue-agents}, 2022.

\bibitem[Helber et~al.(2019)Helber, Bischke, Dengel, and Borth]{helber2019eurosat}
Patrick Helber, Benjamin Bischke, Andreas Dengel, and Damian Borth.
\newblock Eurosat: A novel dataset and deep learning benchmark for land use and land cover classification.
\newblock \emph{Journal of Selected Topics in Applied Earth Observations and Remote Sensing}, 2019.
\newblock URL \url{https://arxiv.org/abs/1709.00029}.

\bibitem[Hou et~al.(2017)Hou, Zhang, and Zhou]{hou2017learning}
Bo-Jian Hou, Lijun Zhang, and Zhi-Hua Zhou.
\newblock Learning with feature evolvable streams.
\newblock \emph{Advances in Neural Information Processing Systems}, 30, 2017.

\bibitem[Huang et~al.(2024)Huang, Ye, Chen, He, Yue, and Ouyang]{Huang2024}
Chenyu Huang, Peng Ye, Tao Chen, Tong He, Xiangyu Yue, and Wanli Ouyang.
\newblock Emr-merging: Tuning-free high-performance model merging.
\newblock \emph{arXiv preprint arXiv:2405.17461}, May 2024.

\bibitem[Huh et~al.(2024)Huh, Cheung, Bernstein, Isola, and Agrawal]{Huh2024}
Minyoung Huh, Brian Cheung, Jeremy Bernstein, Phillip Isola, and Pulkit Agrawal.
\newblock Training neural networks from scratch with parallel low-rank adapters.
\newblock \emph{arXiv preprint arXiv:2402.16828}, February 2024.
\newblock Submitted on 26 Feb 2024, last revised 26 Jul 2024.

\bibitem[Ilharco et~al.(2022)Ilharco, Wortsman, Gadre, Song, Hajishirzi, Kornblith, Farhadi, and Schmidt]{ilharco2022patching}
Gabriel Ilharco, Mitchell Wortsman, Samir~Yitzhak Gadre, Shuran Song, Hannaneh Hajishirzi, Simon Kornblith, Ali Farhadi, and Ludwig Schmidt.
\newblock Patching open-vocabulary models by interpolating weights.
\newblock In \emph{Advances in Neural Information Processing Systems (NeurIPS)}, 2022.
\newblock URL \url{https://arxiv.org/abs/2208.05592}.

\bibitem[Ilharco et~al.(2023)Ilharco, Ribeiro, Wortsman, Gururangan, Schmidt, Hajishirzi, and Farhadi]{ilharco2023editing}
Gabriel Ilharco, Marco~Túlio Ribeiro, Mitchell Wortsman, Suchin Gururangan, Ludwig Schmidt, Hannaneh Hajishirzi, and Ali Farhadi.
\newblock Editing models with task arithmetic.
\newblock In \emph{International Conference on Learning Representations (ICLR)}, 2023.
\newblock URL \url{https://arxiv.org/abs/2110.08207}.

\bibitem[Jacot et~al.(2018)Jacot, Gabriel, and Hongler]{jacot2018neural}
Arthur Jacot, Franck Gabriel, and Cl{\'e}ment Hongler.
\newblock Neural tangent kernel: Convergence and generalization in neural networks.
\newblock In \emph{Advances in Neural Information Processing Systems (NeurIPS)}, 2018.
\newblock URL \url{https://proceedings.neurips.cc/paper_files/paper/2018/file/5a4be1fa34e62bb8a6ec6b91d2462f5a-Paper.pdf}.

\bibitem[Krause et~al.(2013)Krause, Stark, Deng, and Fei-Fei]{krause2013object}
Jonathan Krause, Michael Stark, Jia Deng, and Li~Fei-Fei.
\newblock 3d object representations for fine-grained categorization.
\newblock In \emph{International Conference on Computer Vision Workshops (ICCVw)}, 2013.
\newblock URL \url{https://www.cv-foundation.org/openaccess/content_iccv_workshops_2013/W19/html/Krause_3D_Object_Representations_2013_ICCV_paper.html}.

\bibitem[LeCun(1998)]{lecun1998mnist}
Yann LeCun.
\newblock The mnist database of handwritten digits, 1998.
\newblock URL \url{http://yann.lecun.com/exdb/mnist/}.

\bibitem[Li et~al.(2023)Li, Peng, Zhang, Ding, Hu, and Shen]{li2023deep}
Weishi Li, Yong Peng, Miao Zhang, Liang Ding, Han Hu, and Li~Shen.
\newblock Deep model fusion: A survey, 2023.
\newblock arXiv preprint arXiv:2309.15698.

\bibitem[Liu et~al.(2024)Liu, Golatkar, and Soatto]{Liu2024}
Tian~Yu Liu, Aditya Golatkar, and Stefano Soatto.
\newblock Tangent transformers for composition, privacy and removal.
\newblock In \emph{Proceedings of the International Conference on Learning Representations (ICLR)}, 2024.

\bibitem[Loshchilov \& Hutter(2019)Loshchilov and Hutter]{loshchilov2019decoupled}
Ilya Loshchilov and Frank Hutter.
\newblock Decoupled weight decay regularization.
\newblock In \emph{International Conference on Learning Representations (ICLR)}, 2019.
\newblock URL \url{https://openreview.net/forum?id=Bkg6RiCqY7}.

\bibitem[Lu et~al.(2022)Lu, Welleck, Jiang, Hessel, Qin, West, Ammanabrolu, and Choi]{lu2022quark}
Ximing Lu, Sean Welleck, Liwei Jiang, Jack Hessel, Lianhui Qin, Peter West, Prithviraj Ammanabrolu, and Yejin Choi.
\newblock Quark: Controllable text generation with reinforced unlearning.
\newblock In \emph{Advances in Neural Information Processing Systems (NeurIPS)}, 2022.
\newblock URL \url{https://arxiv.org/abs/2205.13636}.

\bibitem[Matena \& Raffel(2021)Matena and Raffel]{matena2021merging}
Michael Matena and Colin Raffel.
\newblock Merging models with fisher-weighted averaging.
\newblock In \emph{Advances in Neural Information Processing Systems (NeurIPS)}, 2021.
\newblock URL \url{https://arxiv.org/abs/2111.09832}.

\bibitem[McCloskey \& Cohen(1989)McCloskey and Cohen]{mccloskey1989catastrophic}
Michael McCloskey and Neal~J Cohen.
\newblock Catastrophic interference in connectionist networks: The sequential learning problem.
\newblock In \emph{Psychology of Learning and Motivation}. Elsevier, 1989.
\newblock URL \url{https://www.sciencedirect.com/science/article/abs/pii/S0079742108605368}.

\bibitem[Netzer et~al.(2011)Netzer, Wang, Coates, Bissacco, Wu, and Ng]{netzer2011reading}
Yuval Netzer, Tao Wang, Adam Coates, Alessandro Bissacco, Bo~Wu, and Andrew~Y Ng.
\newblock Reading digits in natural images with unsupervised feature learning.
\newblock In \emph{Advances in Neural Information Processing Systems (NeurIPS) Workshops}, 2011.
\newblock URL \url{https://storage.googleapis.com/pub-tools-public-publication-data/pdf/37648.pdf}.

\bibitem[Ortiz-Jimenez et~al.(2024)Ortiz-Jimenez, Favero, and Frossard]{ortiz-jimenez2024task}
G.~Ortiz-Jimenez, A.~Favero, and P.~Frossard.
\newblock Task arithmetic in the tangent space: Improved editing of pre-trained models.
\newblock In \emph{Advances in Neural Information Processing Systems}, volume~36, 2024.

\bibitem[Ortiz-Jiménez et~al.(2021)Ortiz-Jiménez, Modas, Moosavi-Dezfooli, and Frossard]{ortiz-jimenez2021optimism}
Guillermo Ortiz-Jiménez, Apostolos Modas, Seyed-Mohsen Moosavi-Dezfooli, and Pascal Frossard.
\newblock Optimism in the face of adversity: Understanding and improving deep learning through adversarial robustness.
\newblock \emph{Proceedings of the IEEE}, 2021.
\newblock URL \url{https://ieeexplore.ieee.org/document/9348948}.

\bibitem[Ouyang et~al.(2022)Ouyang, Wu, Jiang, Almeida, Wainwright, Mishkin, Zhang, Agarwal, Slama, Ray, et~al.]{ouyang2022training}
Long Ouyang, Jeff Wu, Xu~Jiang, Diogo Almeida, Carroll~L Wainwright, Pamela Mishkin, Chong Zhang, Sandhini Agarwal, Katarina Slama, Alex Ray, et~al.
\newblock Training language models to follow instructions with human feedback, 2022.
\newblock URL \url{https://arxiv.org/abs/2203.02155}.

\bibitem[Radford et~al.(2021)Radford, Kim, Hallacy, Ramesh, Goh, Agarwal, Sastry, Askell, Mishkin, Clark, Krueger, and Sutskever]{radford2021learning}
Alec Radford, Jong~Wook Kim, Chris Hallacy, Aditya Ramesh, Gabriel Goh, Sandhini Agarwal, Girish Sastry, Amanda Askell, Pamela Mishkin, Jack Clark, Gretchen Krueger, and Ilya Sutskever.
\newblock Learning transferable visual models from natural language supervision.
\newblock In \emph{International Conference on Machine Learning (ICML)}, 2021.
\newblock URL \url{https://arxiv.org/abs/2103.00020}.

\bibitem[Ribeiro \& Lundberg(2022)Ribeiro and Lundberg]{ribeiro2022adaptive}
Marco~Tulio Ribeiro and Scott Lundberg.
\newblock Adaptive testing and debugging of nlp models.
\newblock In \emph{Annual Meeting of the Association for Computational Linguistics (ACL)}, 2022.
\newblock URL \url{https://aclanthology.org/2022.acl-long.230/}.

\bibitem[Santurkar et~al.(2021)Santurkar, Tsipras, Elango, Bau, Torralba, and Madry]{santurkar2021editing}
Shibani Santurkar, Dimitris Tsipras, Mahalaxmi Elango, David Bau, Antonio Torralba, and Aleksander Madry.
\newblock Editing a classifier by rewriting its prediction rules.
\newblock In \emph{Advances in Neural Information Processing Systems (NeurIPS)}, 2021.
\newblock URL \url{https://proceedings.neurips.cc/paper_files/paper/2021/file/c46489a2d5a9a9ecfc53b17610926ddd-Paper.pdf}.

\bibitem[Stallkamp et~al.(2011)Stallkamp, Schlipsing, Salmen, and Igel]{stallkamp2011german}
Johannes Stallkamp, Marc Schlipsing, Jan Salmen, and Christian Igel.
\newblock The german traffic sign recognition benchmark: A multi-class classification competition.
\newblock In \emph{International Joint Conference on Neural Networks (IJCNN)}, 2011.
\newblock URL \url{https://ieeexplore.ieee.org/document/6033395}.

\bibitem[Stoica et~al.(2023)Stoica, Bolya, Bjorner, Hearn, and Hoffman]{stoica2023zipit}
George Stoica, Daniel Bolya, Jakob Bjorner, Taylor Hearn, and Judy Hoffman.
\newblock Zipit! merging models from different tasks without training.
\newblock In \emph{International Conference on Learning Representations (ICLR)}, 2023.
\newblock URL \url{https://arxiv.org/abs/2305.03053}.

\bibitem[Tam et~al.(2023)Tam, Bansal, and Raffel]{Tam2023}
Derek Tam, Mohit Bansal, and Colin Raffel.
\newblock Merging by matching models in task subspaces.
\newblock \emph{arXiv preprint arXiv:2312.04339}, December 2023.

\bibitem[Tancik et~al.(2020)Tancik, Srinivasan, Mildenhall, Fridovich-Keil, Raghavan, Singhal, Ramamoorthi, Barron, and Ng]{tancik2020fourier}
Matthew Tancik, Pratul~P. Srinivasan, Ben Mildenhall, Sara Fridovich-Keil, Nithin Raghavan, Utkarsh Singhal, Ravi Ramamoorthi, Jonathan~T. Barron, and Ren Ng.
\newblock Fourier features let networks learn high frequency functions in low dimensional domains.
\newblock In \emph{Advances in Neural Information Processing Systems (NeurIPS)}, 2020.
\newblock URL \url{https://proceedings.neurips.cc/paper/2020/file/55053683268957697aa39fba6f231c68-Paper.pdf}.

\bibitem[Tang et~al.(2023)Tang, Shen, Luo, Zhan, Hu, Du, and Tao]{tang2023parameter}
A.~Tang, L.~Shen, Y.~Luo, Y.~Zhan, H.~Hu, B.~Du, and D.~Tao.
\newblock Parameter-efficient multi-task model fusion with partial linearizeation.
\newblock In \emph{The Twelfth International Conference on Learning Representations}, October 2023.

\bibitem[Wang et~al.(2018)Wang, Singh, Michael, Hill, Levy, and Bowman]{Wang2018GLUE}
Alex Wang, Amanpreet Singh, Julian Michael, Felix Hill, Omer Levy, and Samuel Bowman.
\newblock Glue: A multi-task benchmark and analysis platform for natural language understanding.
\newblock In \emph{Proceedings of the 2018 EMNLP Workshop BlackboxNLP: Analyzing and Interpreting Neural Networks for NLP}, pp.\  353--355, Brussels, Belgium, 2018. Association for Computational Linguistics.
\newblock \doi{10.18653/v1/W18-5446}.
\newblock URL \url{http://aclweb.org/anthology/W18-5446}.

\bibitem[Wortsman et~al.(2022)Wortsman, Ilharco, Gadre, Roelofs, Gontijo-Lopes, Morcos, Namkoong, Farhadi, Carmon, Kornblith, et~al.]{wortsman2022model}
Mitchell Wortsman, Gabriel Ilharco, Samir~Yitzhak Gadre, Rebecca Roelofs, Raphael Gontijo-Lopes, Ari~S Morcos, Hongseok Namkoong, Ali Farhadi, Yair Carmon, Simon Kornblith, et~al.
\newblock Model soups: averaging weights of multiple fine-tuned models improves accuracy without increasing inference time.
\newblock In \emph{International Conference on Machine Learning (ICML)}, 2022.
\newblock URL \url{https://arxiv.org/abs/2203.05482}.

\bibitem[Xiao et~al.(2024)Xiao, Li, Xie, Getzen, Fang, Long, and Su]{xiao2024algorithmic}
Jiancong Xiao, Ziniu Li, Xingyu Xie, Emily Getzen, Cong Fang, Qi~Long, and Weijie~J Su.
\newblock On the algorithmic bias of aligning large language models with rlhf: Preference collapse and matching regularization.
\newblock \emph{arXiv preprint arXiv:2405.16455}, 2024.

\bibitem[Xiao et~al.(2016)Xiao, Ehinger, Hays, Torralba, and Oliva]{xiao2016sun}
Jianxiong Xiao, Krista~A Ehinger, James Hays, Antonio Torralba, and Aude Oliva.
\newblock Sun database: Exploring a large collection of scene categories.
\newblock \emph{International Journal of Computer Vision (IJCV)}, 2016.
\newblock URL \url{https://link.springer.com/article/10.1007/s11263-014-0748-y}.

\bibitem[Yadav et~al.(2023)Yadav, Tam, Choshen, Raffel, and Bansal]{yadav2023resolving}
Prateek Yadav, Derek Tam, Leshem Choshen, Colin Raffel, and Mohit Bansal.
\newblock Resolving interference when merging models.
\newblock In \emph{Advances in Neural Information Processing Systems (NeurIPS)}, 2023.
\newblock URL \url{https://arxiv.org/abs/2306.01708}.

\bibitem[Yang et~al.(2023)Yang, Wang, Shen, Liu, Guo, Wang, and Tao]{yang2023adamerging}
E.~Yang, Z.~Wang, L.~Shen, S.~Liu, G.~Guo, X.~Wang, and D.~Tao.
\newblock Adamerging: Adaptive model merging for multi-task learning.
\newblock In \emph{The Twelfth International Conference on Learning Representations}, October 2023.

\bibitem[Yang et~al.(2024)Yang, Shen, Guo, Wang, Cao, Zhang, and Tao]{Yang2024}
Enneng Yang, Li~Shen, Guibing Guo, Xingwei Wang, Xiaochun Cao, Jie Zhang, and Dacheng Tao.
\newblock Model merging in llms, mllms, and beyond: Methods, theories, applications and opportunities.
\newblock \emph{arXiv preprint arXiv:2405.17461}, May 2024.

\bibitem[Zhang et~al.(2023)Zhang, Chen, Liu, and He]{zhang2023composing}
Jinghan Zhang, Shiqi Chen, Junteng Liu, and Junxian He.
\newblock Composing parameter-efficient modules with arithmetic operations.
\newblock In \emph{Advances in Neural Information Processing Systems (NeurIPS)}, 2023.
\newblock URL \url{https://arxiv.org/abs/2306.14870}.

\bibitem[Zhuang et~al.(2020)Zhuang, Qi, Duan, Xi, Zhu, Zhu, Xiong, and He]{zhuang2020comprehensive}
Fuzhen Zhuang, Zhiyuan Qi, Keyu Duan, Dongbo Xi, Yongchun Zhu, Hengshu Zhu, Hui Xiong, and Qing He.
\newblock A comprehensive survey on transfer learning.
\newblock \emph{Proceedings of the IEEE}, 2020.
\newblock URL \url{https://ieeexplore.ieee.org/document/9134370}.

\end{thebibliography}
\bibliographystyle{iclr2025_conference}
\newpage
\appendix

\section{Experimental Details}
\label{app:experiment}

All our experiments were performed using the same hardware consisting of four 3090 NVIDIA GPUs with 24GB of memory each which can be reproduced in less than 150 GPU hours. The details of each experiment are the following.

\textbf{Datasets.}
We evaluate task arithmetic on a set of popular benchmark datasets from various domains. The dataset collection includes:

\begin{itemize}
    \item \textbf{SVHN} \citep{netzer2011reading}: The Street View House Numbers dataset is a real-world image dataset for developing machine learning and object recognition algorithms with minimal requirements on data preprocessing and formatting.
    \item \textbf{MNIST} \citep{lecun1998mnist}: A database of handwritten digits, with 60,000 training images and 10,000 testing images.
    \item \textbf{EuroSAT} \citep{helber2019eurosat}: A dataset based on Sentinel-2 satellite images covering 13 spectral bands, with 10 classes and a total of 27,000 labeled and geo-referenced images.
    \item \textbf{RESISC45} \citep{cheng2017remote}: The remote sensing image scene classification dataset, consisting of 31,500 images in 45 scene classes.
    \item \textbf{Cars} \citep{krause2013object}: This dataset contains images of cars categorized into various fine-grained classes. It is widely used for fine-grained image classification tasks, providing a rich set of vehicle images for training and evaluation.
    \item \textbf{DTD (Describable Textures Dataset)} \citep{cimpoi2014describing}: This dataset is designed for texture recognition and categorization. It consists of texture images organized into 47 categories, each labeled with attributes describing the texture patterns. It is commonly used to evaluate texture recognition algorithms.
    \item \textbf{SUN397} \citep{xiao2016sun}: The Scene UNderstanding (SUN) dataset is a large-scale dataset for scene recognition, containing 397 categories with a total of over 100,000 images. It is used to evaluate scene understanding models and to benchmark scene classification algorithms.
    \item \textbf{GTSRB (German Traffic Sign Recognition Benchmark)} \citep{stallkamp2011german}: This dataset comprises images of German traffic signs, classified into over 40 categories. It is used to develop and evaluate traffic sign recognition systems, particularly in the context of autonomous driving and intelligent transportation systems.

\end{itemize}

\textbf{Fine-tuning.}
All the fine-tuning experiments follow the same training protocol specified in Ilharco et al. \citep{ilharco2022patching} with minor modifications to the training code to use linearized models when needed. In particular, we fine-tune all datasets starting from the same CLIP pre-trained checkpoint downloaded from the \texttt{open\_clip} repository \citep{Cherti_2023_CVPR}. We fine-tune for 2,000 iterations with a batch size of 128, a learning rate of $10^{-5}$ and a cosine annealing learning rate schedule with 200 warm-up steps and the AdamW optimizer \citep{loshchilov2019decoupled}. As introduced in Ilharco et al. \citep{ilharco2022patching}, during fine-tuning, we freeze the weights of the classification layer obtained by encoding a standard set of zero-shot template prompts for each dataset. Freezing this layer does not harm accuracy and ensures that no additional learnable parameters are introduced during fine-tuning \citep{ilharco2022patching}. We use this exact same protocol to fine-tune the non-linear and linearized models.

\textbf{Tuning of $\alpha$ in Task Arithmetic Benchmarks.}
As in Ilharco et al. \citep{ilharco2022patching}, we use a single coefficient \(\alpha\) to tune the size of the task vectors used to modify the pre-trained models. This is equivalent to setting \(\alpha = \alpha_1 = \ldots = \alpha_T\) in Eq. \ref{eq:disentanglement}. In the task addition benchmarks, after fine-tuning, we evaluate different scaling coefficients \(\alpha \in \{0.0, 0.05, 0.1, \ldots, 1.0\}\) and choose the value that achieves the highest target metric on a small held-out proportion of the training set as specified in Ilharco et al. \citep{ilharco2022patching}. Namely, maximum normalized average accuracy, and minimum target accuracy on each dataset that still retains at least 95\% of the accuracy of the pre-trained model on the control task. The tuning of \(\alpha\) is done independently for non-linear fine-tuning, linearized fine-tuning, and post-hoc linearization.

\textbf{Disentanglement Error.}
To produce the weight disentanglement visualizations of Figure \ref{fig:disentanglement}, we compute the value of \(\xi(\alpha_1, \alpha_2)\) on a \(15 \times 15\) grid of equispaced values in \([-2, 2] \times [-2, 2]\). To estimate the disentanglement error, we use a random subset of 2,048 test points for each dataset.

\section{Further Experimental Results}
We now present additional experiments that expand the findings discussed in the main text.
\begin{figure}
    \centering
    \begin{subfigure}[b]{\textwidth}
        \centering
        \includegraphics[width=0.95\textwidth]{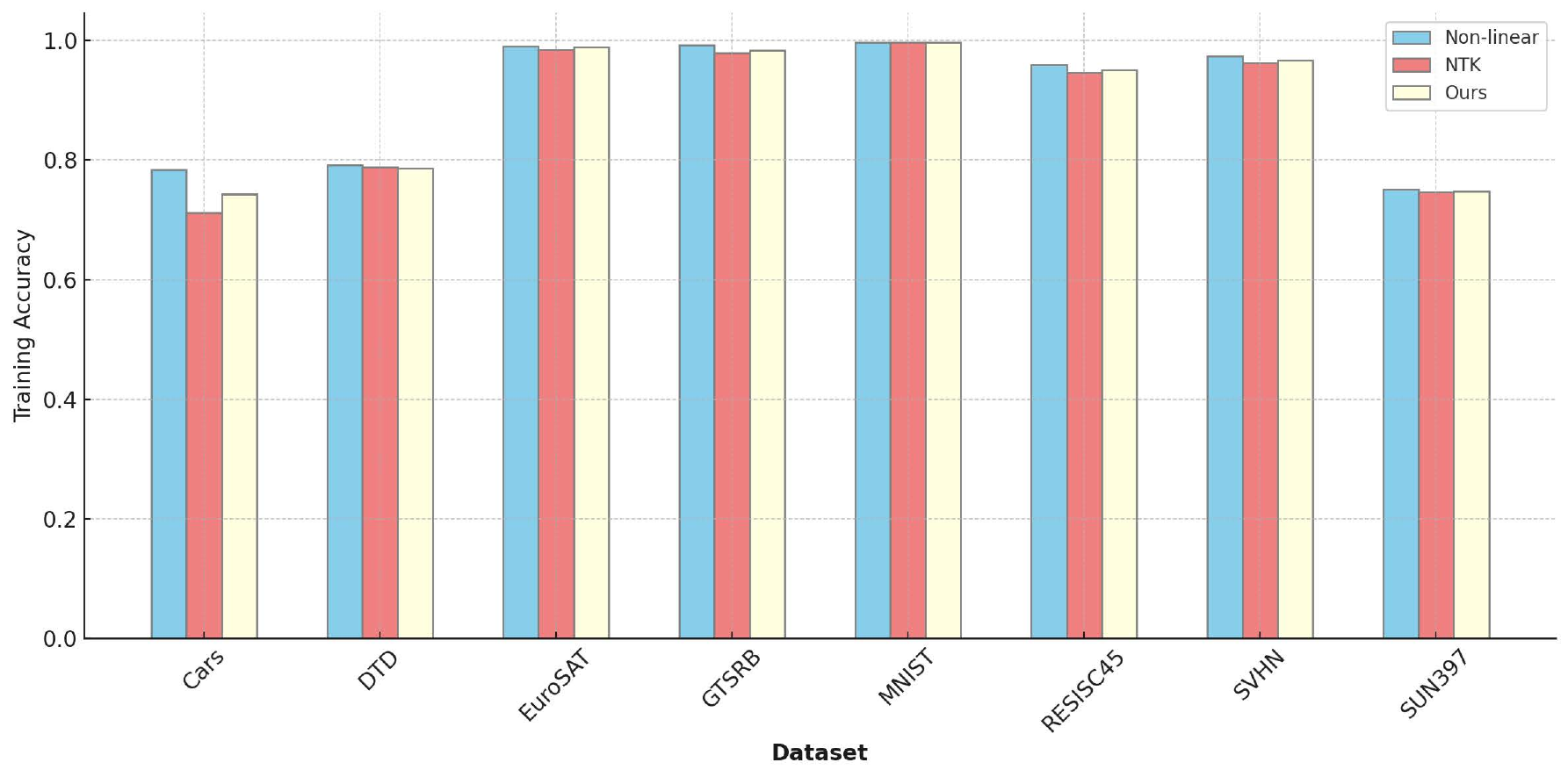}
        \caption{ViT-B/32}
        \label{fig:vit-32}
    \end{subfigure}
    
    
    \begin{subfigure}[b]{\textwidth}
        \centering
        \includegraphics[width=0.95\textwidth]{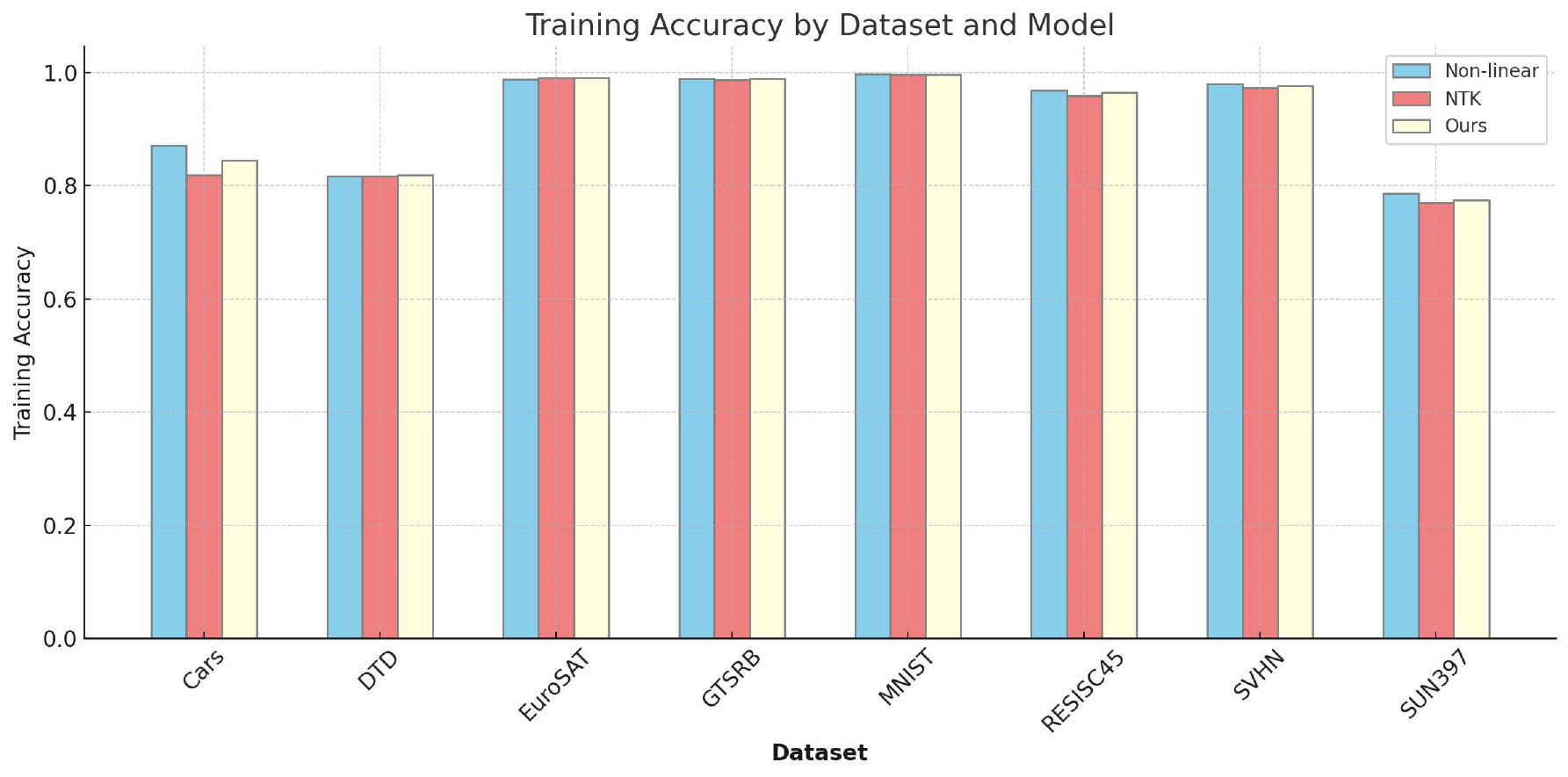}
        \caption{ViT-B/16}
        \label{fig:vit-16}
    \end{subfigure}
    
    
    \begin{subfigure}[b]{\textwidth}
        \centering
        \includegraphics[width=0.95\textwidth]{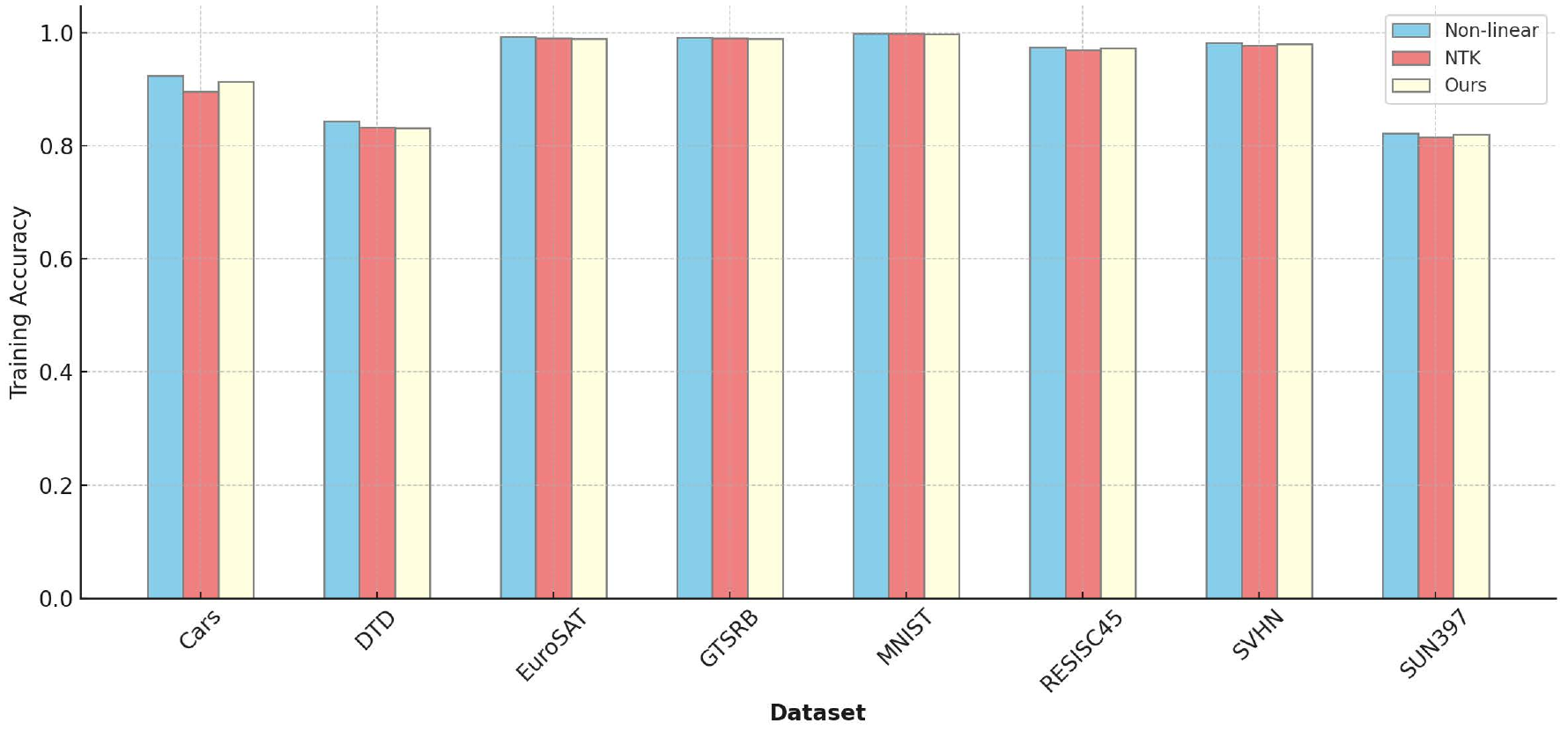}
        \caption{ViT-L/14}
        \label{fig:vit-14}
    \end{subfigure}
    
    \caption{Single-task accuracy of different models obtained using different strategies on each of the tasks.}
    \label{fig:all}
\end{figure}
\subsection{Fine-tuning Accuracy}
\label{app:fine-tuning}
In Figure \ref{fig:all}, we report the single-task accuracy achieved by different CLIP models after fine-tuning with different approaches (referred to as non-linear, NTK linearization, and our method).

\subsection{Weight Disentanglement on Different Task Pairs}
\begin{figure}[H]
    \centering
    \includegraphics[width=\columnwidth]{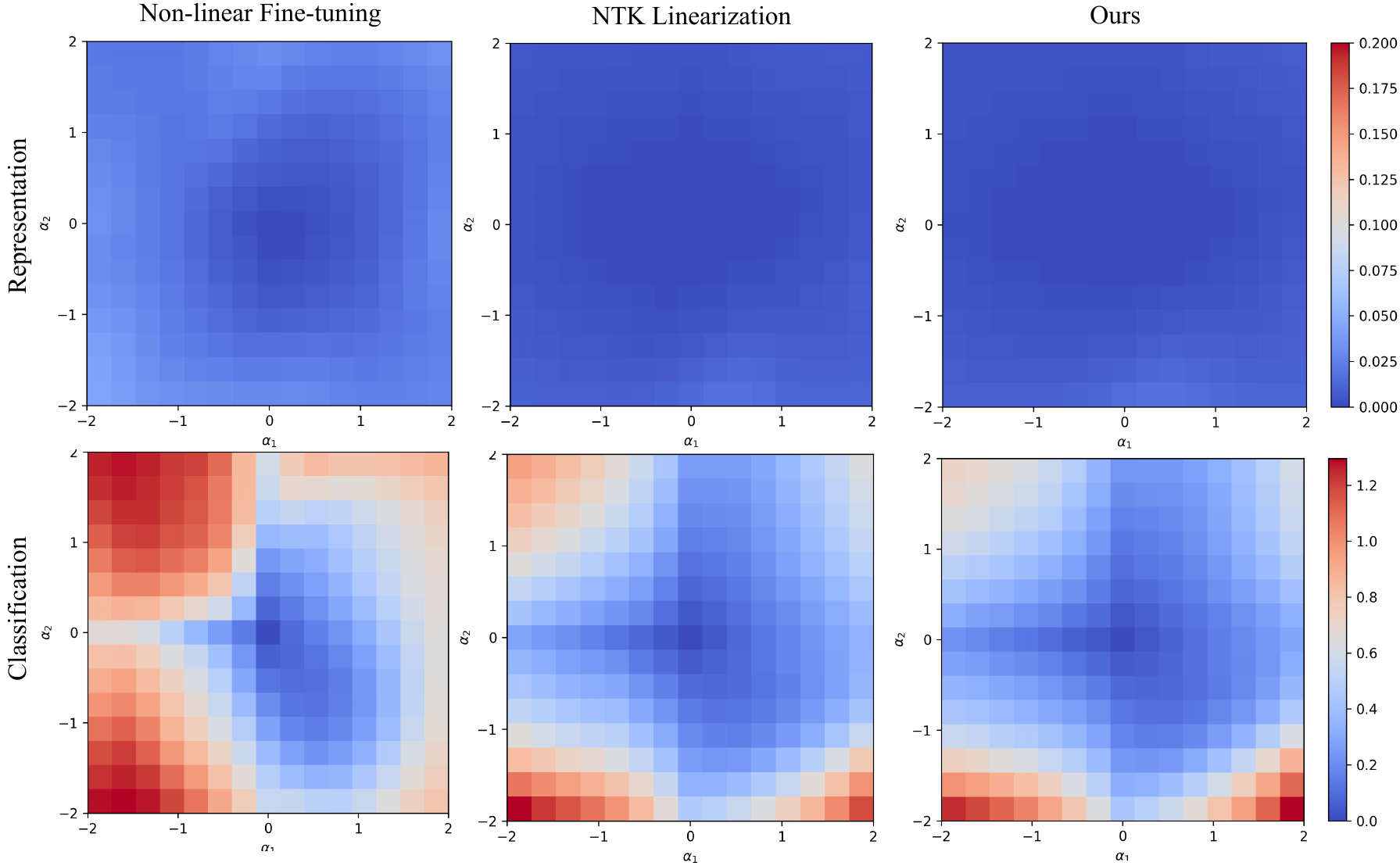}
    \caption{\textbf{Visualization of weight disentanglement.} The heatmaps show the disentanglement error $\xi(\alpha_1, \alpha_2)$ of a single representation module of CLIP ViT-B/32 (top) and a combination of representation module and classification module (bottom) on Cars - RESISC45 task pair. Three fine-tuning paradigms are used from left to right: non-linear fine-tuning, NTK linearization, and ours.
    The light regions denote areas of the weight space where weight disentanglement is stronger.}
    \label{fig:disentanglement_sup}
\end{figure}

In Figure \ref{fig:disentanglement_sup}, we illustrate weight disentanglement on different task pairs.

\begin{figure}
    \centering
    \includegraphics[width=0.8\linewidth]{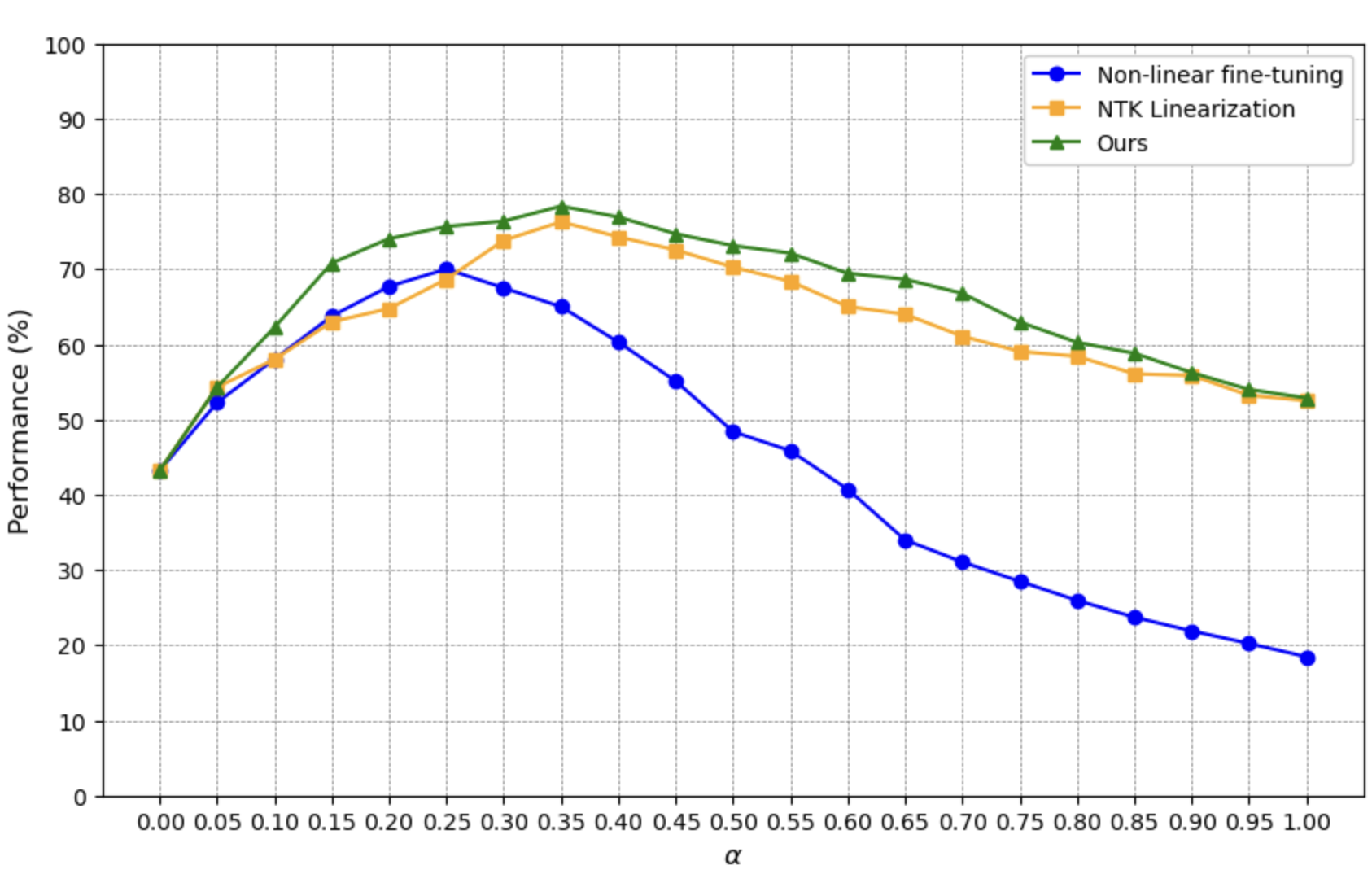}
    \caption{{Performance comparison across different methods with varying $\alpha$ values.}}
    \label{fig:parameter}
\end{figure}

{\subsection{Parameter Sensitivity Analysis}
In our experiments, we replicated the experimental setup used by \cite{ortiz-jimenez2024task} to evaluate the impact of varying $\alpha$ coefficients on model performance. The results are summarized in the Figure \ref{fig:parameter}, which demonstrates that our method exhibits greater robustness across different choices of $\alpha$ compared to both non-linear fine-tuning and NTK linearization.
As illustrated, our method consistently outperforms both non-linear fine-tuning and NTK linearization across a wide range of $\alpha$ values, indicating its robustness in maintaining performance even with varying coefficients.}

\begin{figure}
    \centering
    \includegraphics[width=\columnwidth]{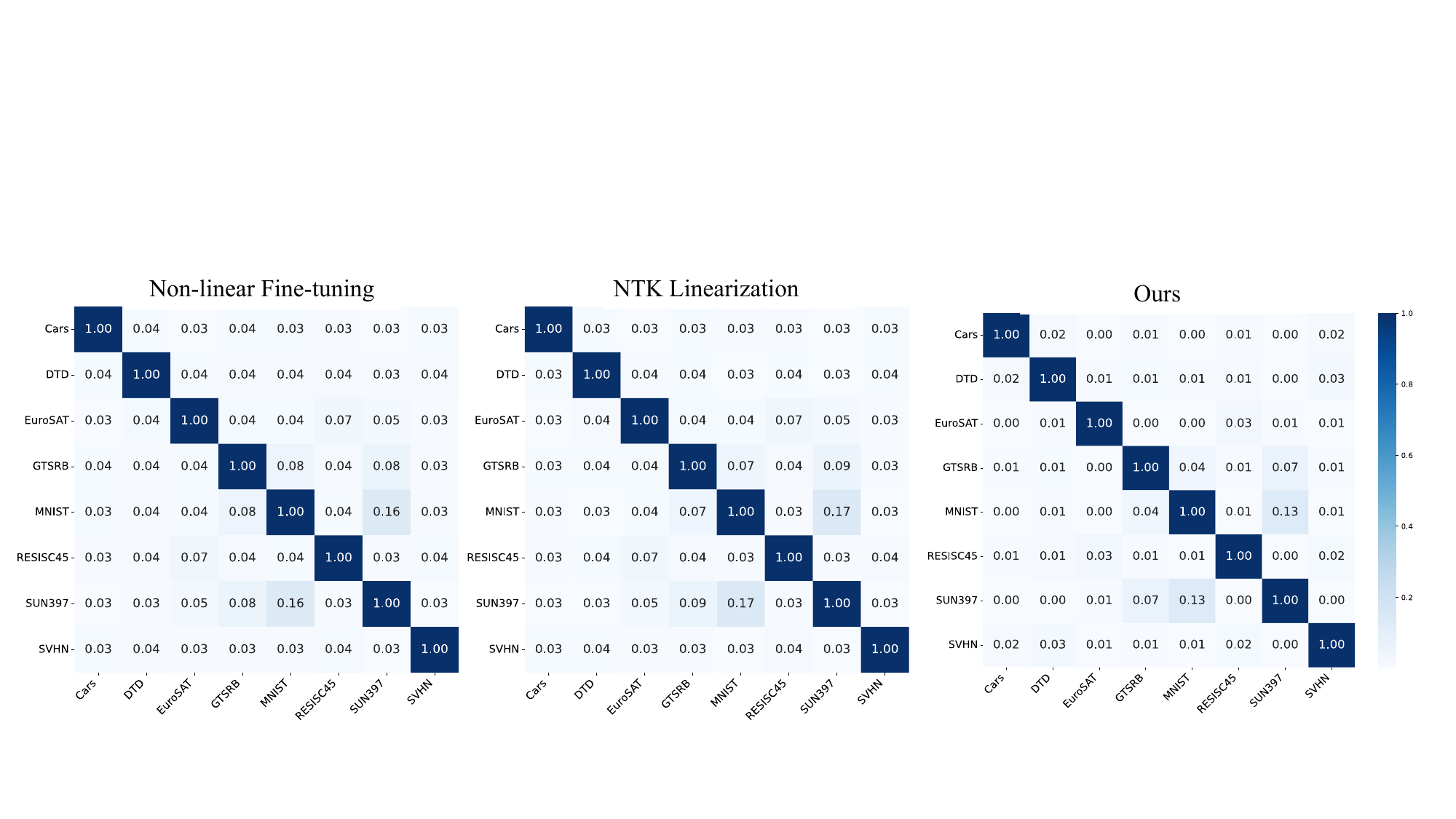}
    \vspace{-0.5cm}
    \caption{{\textbf{Similarity heatmaps.} These figures show heatmaps of the cosine similarity between task vectors from task-specific CLIP models \citep{radford2021learning} fine-tuned on different tasks. Three fine-tuning paradigms from left to right: non-linear fine-tuning, NTK linearization, and Attention modules only fine-tuning (Ours).}}
    \label{fig:cosine similarity}
\vspace{-10pt}
\end{figure}

{\subsection{Similarity between Task Vectors}
Figure \ref{fig:cosine similarity} shows the cosine similarity between task vectors from ViT for three types of fine-tuning (cf. Figure \ref{fig:paradigm}) on image classification tasks. Vectors from attention modules only fine-tuning are closer to orthogonal than those from both non-linear fine-tuning and NTK linearization, indicating that models fine-tuned with full parameters are more independent. This finding aligns with discussions in \citep{ilharco2023editing,tang2023parameter} and is supported by the experimental results in Table \ref{tab:addition}. The experimental details are described in Appendix \ref{app:experiment}.}

\section{Impact Statement}
This paper presents work whose goal is to advance the field of Machine Learning. There are many potential societal consequences of our work, none of which we feel must be specifically highlighted here.

\end{document}